\documentclass{article} 
\usepackage{iclr2025_conference}

\usepackage{amsmath,amsfonts,bm}

\def\eqref#1{equation~\ref{#1}}

\def\1{\bm{1}}

\def\rmD{{\mathbf{D}}}

\def\rmM{{\mathbf{M}}}

\def\rmX{{\mathbf{X}}}

\def\vmu{{\bm{\mu}}}

\def\vh{{\bm{h}}}

\def\vp{{\bm{p}}}
\def\vq{{\bm{q}}}
\def\vr{{\bm{r}}}

\def\vt{{\bm{t}}}
\def\vu{{\bm{u}}}
\def\vv{{\bm{v}}}

\def\vx{{\bm{x}}}

\def\mD{{\bm{D}}}

\def\mI{{\bm{I}}}

\def\mM{{\bm{M}}}

\def\mR{{\bm{R}}}
\def\mS{{\bm{S}}}

\def\mV{{\bm{V}}}
\def\mW{{\bm{W}}}
\def\mX{{\bm{X}}}

\def\mSigma{{\bm{\Sigma}}}

\DeclareMathAlphabet{\mathsfit}{\encodingdefault}{\sfdefault}{m}{sl}
\SetMathAlphabet{\mathsfit}{bold}{\encodingdefault}{\sfdefault}{bx}{n}

\usepackage[pagebackref,colorlinks,breaklinks,citecolor=gray]{hyperref}
\usepackage{url}
\usepackage{times}
\usepackage{graphicx}
\usepackage{comment}
\usepackage{amsmath,amssymb,amsfonts,mathabx,mathbbol}
\usepackage{acronym}
\usepackage{enumitem}
\PassOptionsToPackage{dvipsnames,table}{xcolor}
\usepackage{hyperref}
\usepackage{balance}
\usepackage{xspace}
\usepackage{setspace}
\usepackage[skip=3pt,font=small]{subcaption}
\usepackage[skip=3pt,font=small]{caption}
\usepackage{caption}
\usepackage[misc]{ifsym}
\usepackage[capitalise]{cleveref}
\usepackage{tabularx}
\usepackage{multirow,multirow,array,makecell}
\usepackage{overpic}
\usepackage{wrapfig}
\usepackage{pifont}
\usepackage{url}
\usepackage{booktabs,tabularx,colortbl}       
\usepackage[utf8]{inputenc} 
\usepackage[T1]{fontenc}    
\usepackage{nicefrac}       
\usepackage[nopatch=footnote]{microtype}      
\usepackage[noend,ruled, vlined]{algorithm2e}
\usepackage{algorithmic}
\usepackage{rotating}
\usepackage{adjustbox}
\usepackage{hvfloat}
\makeatletter
\DeclareRobustCommand\onedot{\futurelet\@let@token\@onedot}
\def\@onedot{\ifx\@let@token.\else.\null\fi\xspace}
\def\eg{\emph{e.g}\onedot} 

\def\ie{\emph{i.e}\onedot}

\makeatother

\crefname{algorithm}{Alg.}{Algs.}
\Crefname{algocf}{Algorithm}{Algorithms}
\crefname{section}{Sec.}{Secs.}
\Crefname{section}{Section}{Sections}
\crefname{table}{Tab.}{Tabs.}
\Crefname{table}{Table}{Tables}
\crefname{figure}{Fig.}{Fig.}
\Crefname{figure}{Figure}{Figure}
\definecolor{revision}{RGB}{0,0,255}

\newcommand{\model}{\text{ArtGS}\xspace}
\newcommand{\best}[1]{\cellcolor[HTML]{ffc5c5}{\textbf{#1}}}
\newcommand{\secbest}[1]{\cellcolor[HTML]{ffebd8}{#1}}

\acrodef{3dgs}[3DGS]{3D Gaussian Splatting}
\acrodef{tsdf}[TSDF]{Truncated Signed Distance Function}
\acrodef{fps}[FPS]{Farthest Point Sampling}

\title{Building Interactable Replicas of
Complex \\Articulated Objects via Gaussian Splatting}

\iclrfinalcopy

\author{
\hspace{-9pt} Yu Liu$^{1,2,*,\ddagger}$, Baoxiong Jia$^{2,*}$, Ruijie Lu$^{2,3}$, Junfeng Ni$^{1,2}$, \textbf{Song-Chun Zhu}$^{1,2,3}$, \textbf{Siyuan Huang}$^{2}$
\\
\hspace{-8pt} \small $^1$Tsinghua University~$^2$State Key Laboratory of General Artificial Intelligence, BIGAI~$^3$\small Peking University
}

\iclrfinalcopy 
\begin{document}

\maketitle

\begin{abstract}
Building interactable replicas of articulated objects is a key challenge in computer vision. Existing methods often fail to effectively integrate information across different object states, limiting the accuracy of part-mesh reconstruction and part dynamics modeling, particularly for complex multi-part articulated objects. We introduce \model, a novel approach that leverages 3D Gaussians as a flexible and efficient representation to address these issues. Our method incorporates canonical Gaussians with coarse-to-fine initialization and updates for aligning articulated part information across different object states, and employs a skinning-inspired part dynamics modeling module to improve both part-mesh reconstruction and articulation learning. Extensive experiments on both synthetic and real-world datasets, including a new benchmark for complex multi-part objects, demonstrate that \model achieves state-of-the-art performance in joint parameter estimation and part mesh reconstruction. Our approach significantly improves reconstruction quality and efficiency, especially for multi-part articulated objects. Additionally, we provide comprehensive analyses of our design choices, validating the effectiveness of each component to highlight potential areas for future improvement. Our work is made publicly available at: \url{https://articulate-gs.github.io}.

\let\thefootnote\relax\footnote{$^*$Equal contribution. $^{\ddagger}$Work done as an intern at BIGAI.}

\end{abstract}

\section{Introduction}
\label{sec:intro}
Articulated objects, central to everyday human-environment interactions, have become a key focus in computer vision research~\citep{yang2023reconstructing,weng2024neural,luo2024physpart,liu2024cage,deng2024articulate}. Accurately reconstructing real-world scenes~\citep{chen2024single,ni2024phyrecon,lu2024movis} and creating interactable digital replicas of these objects are essential for various applications, including scene understanding~\citep{jia2024sceneverse, huang2024embodied, zhu2024unifying,linghu2024multi} and robotics learning~\citep{liu2022akb,geng2023gapartnet,geng2023partmanip,gong2023arnold,yang2024physcene,zhao2024tac,lu2024manigaussian}. By building high-fidelity digital twins of articulated objects, we bridge the gap between synthetic and real-world scenarios, thus facilitating the sim-to-real transfer of robotic systems~\citep{torne2024reconciling,kerr2024rsrd}. As we advance towards more sophisticated robotic systems and immersive virtual environments, there is a growing need for improved and efficient modeling techniques for the reconstruction of articulated objects.



The problem of reconstructing articulated objects has been extensively studied~\citep{jiayi2023paris,liu2023building,weng2024neural,deng2024articulate,yang2023reconstructing}, with a key challenge being the learning of object geometry when only partial views of the object are available at any given state. To accurately reconstruct object parts (\eg, a closed drawer), it is essential to integrate observations from multiple object states during interactions (\eg, the opening process of the drawer). This necessitates the simultaneous learning and alignment of fine-grained object parts across different states, which must be achieved jointly during the reconstruction of object geometries. Such a requirement presents significant challenges in object modeling, especially for complex everyday articulated objects that often consist of multiple interactable parts. Additionally, uncertainties in object geometry reconstruction introduce further challenges in modeling articulation, as errors in geometry modeling can result in inaccurate learning of articulation parameters. These challenges highlight the need for improved models that handle the complexities of multi-part articulated objects.

Recent approaches attempt to address these challenges using part priors from pre-trained models. These models provide either part segmentation masks via models like SAM~\citep{kirillov2023segment}, or 2D pixel correspondences for aligning pixels across states~\citep{sun2021loftr}. However, these methods rely heavily on priors from pre-trained models, often using single-state inputs and neglecting critical motion information~\citep{mandi2024real2code} and struggling with the complexity of multi-part objects when accurately matching pixels across states becomes difficult~\citep{weng2024neural}. These limitations result in unstable and inconsistent learning of object parts, posing significant challenges to the joint learning of part motion and geometry.

To address these challenges, we propose \model, which introduces several key innovations for handling complex multi-part articulated objects. Specifically, we adopt the commonly used two-state setting for learning articulated objects, as established in prior works~\citep{jiayi2023paris, weng2024neural}. Central to our approach is the use of 3D Gaussians~\citep{kerbl20233d} as the foundational representation, chosen for their ability to explicitly maintain spatial information while offering efficiency and high reconstruction quality. To effectively model object dynamics and integrate information across multiple object states, we employ canonical Gaussians with a carefully designed coarse-to-fine initialization and update scheme. These Gaussians act as a bridge between different input object states, enabling accurate deformation modeling that improves both mesh reconstruction and articulation learning. Building on the canonical Gaussians, we draw inspiration from Gaussian skinning~\citep{song2024reacto} and introduce a center-based clustering module for part and dynamics learning. This approach leverages motion priors of Gaussians, which are summarized during the learning process, serving as a guide to better align object parts between states and improve articulation learning.
These designs allow our method to achieve state-of-the-art performance in joint parameter estimation and part mesh reconstruction, excelling on both existing benchmarks and our newly curated complex multi-part articulated object reconstruction benchmark. Our approach outperforms existing methods in both synthetic and real-world scenarios, with significant improvements in axis modeling and overall efficiency. Through extensive experiments, we demonstrate the effectiveness of our model in efficiently delivering high-quality reconstruction of complex multi-part articulated objects. We also provide comprehensive analyses of our design choices, highlighting the critical role of these modules and identifying areas for future improvement. 

\paragraph{Contributions} Our main contributions of this work can be summarized as follows:

\begin{itemize}[leftmargin=*,nolistsep,noitemsep]
 \item We propose \model, a novel and efficient method for articulated object reconstruction that achieves state-of-the-art performance, particularly for complex multi-part objects.
 \item We introduce coarse-to-fine canonical Gaussian initialization and skinning-inspired part dynamics modeling with self-guided motion priors to improve object part and articulation learning, effectively addressing the limitations of existing methods in using object motion information.
 \item We conduct extensive experiments on both synthetic and real-world articulated objects, demonstrating the effectiveness, efficiency, scalability, and robustness of our approach. We also provide comprehensive ablation studies to validate our designs and highlight areas for future improvement.
\end{itemize}

\section{Related Work}
\label{sec:related_work}
\paragraph{Dynamic Gaussian Modeling}
\label{sec:related_work:dynamic_gs}
Recent advancements have shown the potential of Gaussian Splatting~\citep{kerbl20233d} for 4D reconstruction~\citep{jung2023deformable, katsumata2023efficient,wu20244d,luiten2024dynamic,li2024spacetime,lu20243d, lei2024gart,guo2024motion,qian20243dgs,bae2024per,wan2024template}. A central focus of these efforts is the deformation modeling of 3D Gaussians. While effective for dynamics capturing, most approaches learn transformations implicitly, limiting their capability for controllable dynamics modeling. To address this issue, recent studies use superpoints~\citep{huang2024sc,wan2024superpoint} for improved dynamics modeling and control. However, as superpoint learning is based primarily on rendering without considering object physics, these methods fail to reliably capture accurate physical parameters (\eg, joints and axes). Another line of works~\citep{xie2024physgaussian,jiang2024vr} introduce controllable Gaussians by integrating physics-based modeling for graphics simulations. These models require intricate priors of objects (\eg, material properties), making them impractical for reconstructing everyday articulated objects. To overcome these challenges, our work combines the explicit 3D Gaussian modeling with articulation modeling, enabling efficient and high-quality reconstruction with precise articulation parameter estimation for more practical digital-twin construction of articulated objects.

\paragraph{Articulation Parameter Estimation}
\label{sec:related_work:artmodel}
Estimating joint articulation parameters for articulated objects has been extensively studied, with approaches broadly categorized into two main categories. First, prediction-based methods estimate joint parameters from sensory inputs of different object configurations \citep{huang2014occlusion,katz2013interactive} or use end-to-end models \citep{hu2017learning,yi2018deep,li2020category,wang2019shape2motion,sun2023opdmulti,liu2022toward,weng2021captra,sturm2011probabilistic,chu2023command,martin2016integrated,liu2023self,gadre2021act,mo2021where2act,jain2021screwnet,yan2020rpm,lei2023nap} to predict part segmentation, kinematic structure, as well as joint parameters. Second, reconstruction-based methods optimize articulation parameters by reconstructing multi-view images or videos~\citep{wei2022self,tseng2022cla,mu2021sdf,lewis2022narf22,jiayi2023paris,lei2024gart,deng2024articulate,swaminathan2024leia,noguchi2022watch,zhang2021strobenet,pillai2015learning,liu2023building}.
Most of these methods treat articulation parameter estimation as a separate task, without generating high-quality, interactable part-mesh reconstructions. \model aims to address this gap by integrating part-mesh reconstruction and articulation parameter estimation, enabling the creation of high-quality, interactable replicas.

\paragraph{Articulated Object Reconstruction}
\label{sec:related_work:reconstruction}
Articulated object reconstruction, differing from human and animal motion modeling~\citep{joo2018total,loper2023smpl,mihajlovic2021leap,noguchi2021neural,yang2021viser,yang2021lasr,romero2022embodied,zuffi20173d,yang2024attrihuman,xu2020ghum,tan2023distilling, yang2022banmo,yang2023ppr,song2023moda,yang2023reconstructing,song2023total}, focus on the piece-wise rigidity of each part, requiring both part-level geometry reconstruction and joint articulation parameter estimation. While end-to-end models predict joint parameters and segment object parts from single-stage~\citep{heppert2023carto, wei2022self,kawana2021unsupervised} or interaction observations\citep{jiang2022ditto, ma2023sim2real, nie2022structure, hsu2023ditto}, they struggle to generalize to unseen objects. Per-object optimization approaches~ \citep{jiayi2023paris,liu2023building,weng2024neural,deng2024articulate,swaminathan2024leia}, using multi-state observations for articulation modeling, offer better adaptability to unknown objects but face scaling issues of multiple joints. Methods like DTA~\citep{weng2024neural} attempt to handle multi-part objects but still struggle with those having more than three movable parts. We address the reliability, flexibility, and scalability issues of previous works with our canonical Gaussian design and skinning-inspired part dynamics modeling, achieving higher accuracy, robustness, and efficiency for articulated object reconstruction.


\section{Preliminaries}
\label{sec:prel}
\paragraph{3D Gaussian Splatting} \ac{3dgs} represents a static 3D scene using 3D Gaussians~\citep{kerbl20233d}. Each Gaussian $G_i$ is associated with a center $\vmu_i$, covariance matrix $\mSigma_i$, opacity $\sigma_i$ and spherical harmonics coefficients $\vh_i$. The final opacity of a 3D Gaussian at a spatial point $\vx$ can be calculated as: 
\begin{equation}
 \begin{aligned}
 \alpha_i(\vx)= \sigma_i \exp\left(-\frac{1}{2}(\vx-\vmu_i)^T\mSigma_i^{-1}(\vx - \vmu_i)\right), && \text{where} && \mSigma_i = \mR_i\mS_i\mS_i^T\mR_i^T.
 \end{aligned}
 \label{eq:gs_opacity}
\end{equation}
As the physical meaning of a covariance matrix is only valid if it is positive semi-definite, we decompose the covariance matrix $\mSigma_i$ following~\cref{eq:gs_opacity} into a scaling diagonal matrix $\mS_i$ and a rotation matrix $\mR_i$ parameterized by a quaternion $\vr_i$. A scene is then described with a collection of such Gaussians $\mathcal{G} = \{G_i:\vmu_i, \vr_i, {\bm{s}}_i, \sigma_i, \vh_i\}_{i=1}^{N}$. We render an image $\mI$ and optionally its depth image $\mD$ from the 3D scene $\mathcal{G}$ by projecting each Gaussian onto the 2D image plane and aggregating them using $\alpha$-blending:
\begin{equation}
 \begin{aligned}
 \mI = \sum_{i=1}^{N}T_i\alpha_i^{\text{2D}}\mathcal{SH}(\vh_i, \vv_i), && \mD=\sum_{i=1}^N T_i\alpha_i^{\text{2D}}d_i, && \text{where} && T_i = \prod_{j=1}^{i-1}(1 - \alpha_j^{\text{2D}}).
 \end{aligned}
 \label{eq:gs_rendering}
\end{equation}
$\alpha_i^{\text{2D}}$ is a 2D version of~\cref{eq:gs_opacity}, with $\vmu_i$, $\mSigma_i$, $\vx$ replaced by the projected $\vmu_i^{\text{2D}}$, $\mSigma_i^{\text{2D}}$, and the pixel coordinate $\vu$. $\mathcal{SH}(\cdot)$ is the spherical harmonic function, $\vv_i$ is the view direction from the camera to $\vmu_i$, $d_i$ is the depth of the $i$-th Gaussian. Given $N_v$ input view images $\{\bar{\mI}_i, \bar{\mD}_i\}_{i=1}^{N_v}$, \ac{3dgs} learns Gaussians $\mathcal{G}$ with:
\begin{equation}
\begin{aligned}
\mathcal{L}_{\text{render}} = (1-\lambda_{\text{SSIM}})\mathcal{L}_I + \lambda_{\text{SSIM}}\mathcal{L}_{\text{D-SSIM}} + \mathcal{L}_D,
\end{aligned}
\label{eq:rendering_loss}
\end{equation}
where $\mathcal{L}_I = ||\mI - \bar{\mI}||_1$ is the L1-loss, $\mathcal{L}_{\text{D-SSIM}}$ is the D-SSIM loss~\citep{kerbl20233d}, $\lambda_{\text{SSIM}}$ is the weight of D-SSIM loss, and $\mathcal{L}_D = \log\left(1 + ||\mD - \bar{\mD}||_1\right)$ is the optional depth supervision. 

\paragraph{Mesh Extraction from Gaussians} To extract meshes from Gaussian splats $\mathcal{G}$, we can render depth maps and utilize \ac{tsdf} to fuse the reconstructed depth maps, and extract the object mesh $\mathcal{M}$ with marching cubes~\citep{huang20242d}. This process can be done with Open3D~\citep{zhou2018open3d} with proper choice of voxel size and truncated threshold.

\section{Method}
\label{sec:method}
Given $N_v$ RGB-D images of an unknown articulated object $\{\bar{\mI}^t_i, \bar{\mD}^t_i\}_{i=1}^{N_v}$ at two joint states $t\in\{0,1\}$, we aim to reconstruct its part-level meshes $\mathcal{M}$ and joint articulation parameters $\Psi$. We define a set of learnable canonical Gaussians $\mathcal{G}^c$ which can be transformed into joint state Gaussians $\mathcal{G}^t$ via a per-Gaussian SE(3) transformation $T^{c\to t}$, parameterized by $\Psi$. Formally,
\begin{equation}
 \begin{aligned}
 \mathcal{G}^t = T^{c\to t}\cdot\mathcal{G}_c \quad \text{and} \quad \mathcal{G}^{c} = (T^{c\to t})^{-1}\cdot\mathcal{G}^t \quad \text{for} \quad t\in\{0,1\}. \\
 \end{aligned}
 \label{eq:motion_model}
\end{equation}
We impose the continuity of motion between the joint states by setting the canonical Gaussians $\mathcal{G}^{c}$ at the mid-state ($c$ : $t$ = 0.5), enforcing that $T^{c\to0}=(T^{c\to1})^{-1}$. This simplifies the articulation learning and connects the two input joint states through the canonical Gaussians $\mathcal{G}^c$, solving potential issues of occlusion and misinformation when reconstructing object meshes separately on the two joint states.

\begin{figure}[t!]
 \centering
 \resizebox{\linewidth}{!}{\includegraphics[width=\linewidth]{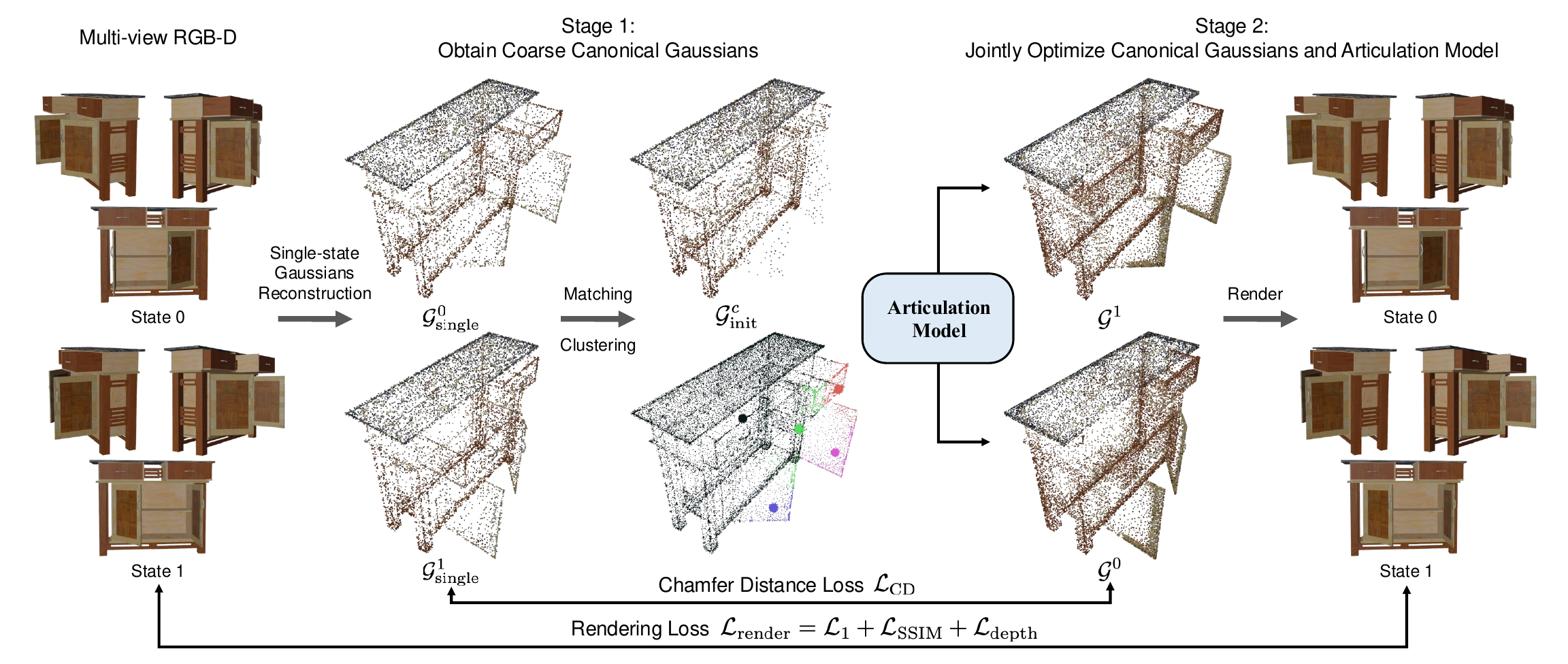}}
 \caption{\textbf{The overview of \model.} Our method is divided into two stages: (i) obtaining coarse canonical Gaussians $\mathcal{G}^c_{\text{init}}$ by matching the Gaussians $\mathcal{G}^0_{\text{single}}$ and $\mathcal{G}^1_{\text{single}}$ trained with each single-state individually and initializing the part assignment module with clustered centers, (ii) jointly optimizing canonical Gaussians $\mathcal{G}^c$ and articulation model (including the articulation parameters $\Psi$ and the part assignment module in \cref{sec:method:skinning}).}
 \label{fig:overview}
\end{figure}

Using this motion model, we leverage multi-view RGB-D images from the two input states to learn both the canonical Gaussian $\mathcal{G}^c$, the transformation $T^{c\to1}$ or equivalently the joint parameters $\Psi$, and extract object meshes $\mathcal{M}^t$ for different joint states following~\cref{sec:prel}. An overview of \model is presented in~\cref{fig:overview}, with details on key designs provided in the following sections.


\subsection{Coarse-to-Fine Canonical Gaussian Initialization with Motion Analysis}
\label{sec:method:canonical}
The initialization of the canonical Gaussians $\mathcal{G}^c$ is crucial for articulation learning. A good initialization leverages the consistency between input joint states, improving mesh reconstruction and articulation modeling. In contrast, a random initialization leads to undesirable local minima, adversely affecting the learning process (see in~\cref{fig:ablation}). To tackle this issue, we propose a coarse-to-fine strategy for the canonical Gaussian initialization, incorporating preliminary motion information from the two input joint states to enhance subsequent articulation modeling.

\paragraph{Coarse Initialization by Matching Single-state Gaussians} 
In this phase, we first separately train two sets of single-state Gaussians $\mathcal{G}^t_{\text{single}}$ with input multi-view images following~\cref{eq:rendering_loss}. We then apply Hungarian Matching to obtain matched Gaussian pairs between $\mathcal{G}_{\text{single}}^0$ and $\mathcal{G}_{\text{single}}^1$, based on the distance between Gaussian centers. We take the mean of each pair of matched Gaussians as the coarse canonical Gaussian initialization $\mathcal{G}_{\text{coarse}}^c$. To reduce the significant computation time associated with matching a large number of Gaussians, we use \ac{fps} to downsample the learned single-state Gaussians to a set of 5K Gaussians prior to matching. 


\paragraph{Initialization Refinement with Motion Analysis} 
To support geometry reconstruction and articulation modeling, relying solely on 5K matched coarse Gaussians alone is insufficient. Therefore, we refine the coarse initialization $\mathcal{G}_{\text{coarse}}^c$ guided by the motion information of object parts.
Intuitively, single-state Gaussians, $\mathcal{G}_{\text{single}}^{0}$ and $\mathcal{G}_{\text{single}}^{1}$, should exhibit consistency for static object parts discrepancies for movable parts, \ie, the static parts of these Gaussians are well-learned. Based on this insight, we refine the set of coarse canonical Gaussians $\mathcal{G}_{\text{coarse}}^c$ by including Gaussians corresponding to static parts, allowing more focused learning of movable parts during articulation modeling. In practice, we classify each Gaussian $G_i$ in a joint state $t$ as static or dynamic by calculating its minimum Chamfer Distance to all Gaussians in the opposite state $\bar{t}$:

\begin{equation}
\text{CD}_i^{t\to \bar{t}} = \min_{j}||\vmu^t_i - \vmu^{\bar{t}}_j||_2,\quad G_i \in \mathcal{G}_{\text{single}}^t, G_j\in\mathcal{G}_{\text{single}}^{\bar{t}}\quad\text{and}\quad \text{CD}^{t\to\bar{t}}=\underset{i}{\mathrm{Mean}}\left(\text{CD}_i^{t\to\bar{t}}\right).
\label{eq:mobility_identification}
\end{equation}

If the distance $\text{CD}_i^{t\to\bar{t}}$ exceeds a threshold $\epsilon_{\text{static}}$, $G_i$ is classified as dynamic; otherwise it is static. To determine which state, $t$ or $\bar{t}$, contains more motion information, we compare the mean distance $\text{CD}^{t\to\bar{t}}$ of all Gaussians in state $t$ following~\cref{eq:mobility_identification} and classify the higher state as the more motion informative state. For instance, a cabinet with open drawers provides clearer identification of movable parts than one with closed drawers. With this information, we add the static Gaussians from the more motion informative state to refine $\mathcal{G}_{\text{coarse}}^c$ into the final initialization of the canonical Gaussian $\mathcal{G}^c_{\text{init}}$.

\subsection{Part Discovery for Articulation Modeling}
\label{sec:method:skinning}
Following~\cref{eq:motion_model}, we use a part-based formulation for articulation modeling. Specifically, given the number of parts $K$, we aim to decompose the Gaussians into $K$ parts and learn the articulation paramerters $\Psi = \{T_{k}^{c\to 1}\}_{k=1}^{K}$. In contrast to existing works that leverage prior information for part discovery~\citep{mandi2024real2code,weng2024neural}, we discover parts in an unsupervised manner during learning.


\paragraph{Center-based Part Modeling and Assignment}
Given input canonical Gaussians $\mathcal{G}^c = \{G_i\}_{i=1}^{N}$, our objective is to compute part-level masks $\rmM\in\mathbb{R}^{N\times K}$ that assign each Gaussian $\mathcal{G}_i$ to a specific part. A common approach to generating these assignment masks is through unsupervised segmentation modules using MLPs or slot-attention~\citep{locatello2020sa,jia2023improving,liu2025slotlifter}. However, these models implicitly segment parts and fail to leverage the explicit spatial and dynamic information present in 3D Gaussians. We observe that such methods struggle with parts that exhibit similar motion patterns, leading to incorrect assignments. To address this issue, we adopt a center-based part modeling approach that explicitly utilizes spatial information, inspired by sparse control points from SC-GS~\citep{huang2024sc} and quasi-rigid blend skinning in REACTO~\citep{song2024reacto}. Specifically, we define $K$ learnable centers $C_k = (\vp_k, \mV_k, {\bm \lambda}_k)$ with center location $\vp_k\in\mathbb{R}^3$, rotation matrix $\mV_k\in\mathbb{R}^{3\times 3}$, and scale vector ${\bm \lambda}_k\in\mathbb{R}^3$. 
For a given Gaussian $G_i \in \mathcal{G}^c$, we compute the Mahalanobis distance $\mD_{ik}$ between $G_i$ and center $C_k$ as:
\begin{equation}
 \rmX^k_i = \frac{[\mV_k(\vmu^c_i-\vp_k)]}{{\bm\lambda}_k} \quad
 \rmD_{i}^k = (\rmX^k_i)^T\cdot\rmX^k_i\quad\text{and}\quad\bm{M} = \mathrm{GumbelSoftmax}\left(\frac{-\mD + \mW_{\Delta}}{\tau}\right)
 \label{eq:part_assignment}
\end{equation}
where $\mD_{i}^k$ is the distance matrix for part assignment. One challenge of using the distance matrix for part assignment is identifying sharp boundaries when two parts overlap spatially (\eg, in the case of a closed drawer). To improve boundary identification, we introduce a residual term $\mW_\Delta = \mathrm{MLP}(\vmu, \mX, \mD)$, predicted by a shallow MLP that concatenates the absolute position of each Gaussian and the distance matrix $\mD$ as input. This residual is added to the original distance matrix $\mD$ to refine the part assignment mask following \cref{eq:part_assignment}. Notably, we use Gumbel Softmax to ensure that each Gaussian is assigned to only one part, which simplifies the optimization of joint parameters. Detailed implementation can be found in~\cref{app:imp:assignment}.

\paragraph{Center Initialization by Clustering Coarse Gaussians}
We empirically find that the initialization of centers $\vp_k$ and scale ${\bm \lambda}_k$ have great impacts on the correctness of part discovery in later learning process (see~\cref{fig:ablation}). Therefore, similar to the canonical Gaussian initialization described in~\cref{sec:method:canonical}, we utilize the motion type of each joint as additional information for providing good initializations of part centers. Specifically, we select the input joint state with more motion information to identify static and dynamic parts. For static parts, we take the mean of the Gaussians as the part center. For movable parts, we do spectral clustering on the positions of movable Gaussians ($K-1$ clusters) and take the mean of each cluster for part center initialization. We use the distance from the farthest point to the center of each cluster as the initial scale.


\subsection{Self-guided Articulation Type and Parameter Learning}\label{sec:method:dual_quaternion}
After obtaining object part representations, we define the per-part articulation parameters via dual-quaternions. Formally, the joints articulation parameters $\Psi = \{T_k^{c\to 1}\}_{k=1}^{K} = \{\vq_k^{c\to1}:({\vq}_{k,r}, \vq_{k, d})\}_{k=1}^K$, where $\vq_{k,r}$ and ${\vq}_{k,d}$ are the real and dual part of the dual-quaternion that determine the rotation and translation of the joint transformation respectively. For notational simplicity, we use $\vq_k^t$ for $\vq_k^{c\rightarrow t}$ in the following texts. With the mid-state assumption in~\cref{sec:method}, we have $\vq_{k}^0 = (\vq_k^{1})^{-1}$ is the inverse of dual-quaternion $\vq_{k}^1$. Given object masks $\mM$ obtained in~\cref{sec:method:skinning}, the per-gaussian dual-quaternion $\vq_i$ for Gaussian $G_i \in \mathcal{G}^c$ is given by:
\begin{equation}
\begin{aligned}
\vq_i^t =(\sum_{k=1}^K \mM_{ik}\cdot {\vq^t_{k,r}},\sum_{k=1}^K \mM_{ik}\cdot {\vq^t_{k,d}}). 
\end{aligned}
\end{equation}
where $(\vq_k^1)^{-1}$ is the inverse of dual-quaternion $\vq_k^{1}$ and $\mM_{ik}$ denotes the probability of Gaussian $i$ belongs to part $k$.
With the per-gaussian transformation given $\vq_{i}^{t}$, we transform the canonical Gaussian $\mathcal{G}^c$ to get the two joint state Gaussians $\mathcal{G}^t$ with:
\begin{equation}
\begin{aligned}
\vmu_i^t = \mR_i^{c\rightarrow t}\cdot \vmu_i^c + \vt_i^{c\rightarrow t}, \quad \vr_i^t = {\vq}_{i,r}^{t} \otimes \vr_i^c,\\
\end{aligned}
\label{eq:articulation_modeling}
\end{equation}
where $\mR_i^{c\rightarrow t}$ and $\vt_i^{c\rightarrow t}$ is the per-gaussian rotation matrix and translation vector derived from $\vq_i^{t}$, and $\otimes$ denotes quaternion multiplication operation. We assume that the scale ${\bm s}_i$ and opacity $\sigma_i$ of the Gaussian $G_i$ remains consistent under transformation.

To enhance the learning of articulation parameters, we adopt a warm-up strategy for predicting the joint type of each part. During the warm-up stage, we optimize the articulation parameters $\Psi=\{\vq_k^{1}\}_{k=1}^{K}$ without any constraints. Next, we develop a heuristic for joint type prediction based on the learned rotation ${\vq_{k,r}}$. Specifically, we classify the joint as revolute if the rotation degree of ${\vq_{k,r}}$ exceeds a threshold $\epsilon_{\text{revol}}$, and otherwise prismatic. With predicted joint types, we constrain the joint transformation for each part. Specifically, we manually set the rotation quaternion $\vq_{k,r}$ of prismatic joints as identity quaternion. This operation allows the model to focus on optimizing the translation term $\vq_{k,d}$ of the prismatic joint, thereby obtaining a more accurate estimate of the joint parameters. 
\subsection{Optimization}
We train our model using the rendering loss with depth supervision $\mathcal{L}_{\text{render}}$ described in~\cref{sec:prel} on the reconstructed $\mathcal{G}^t$ for the two joint states as discussed in~\cref{sec:method:dual_quaternion}. To reduce the chances of learning artifacts during update, we use the single-state reconstructed Gaussians $\mathcal{G}_\text{single}^t$ as an additional supervision:
\begin{equation}
\label{eq:cd_loss}
 \mathcal{L}_{\text{CD}}=\frac{1}{N}\sum_{i=1}^{N}\min_{j}||\vmu^t_i - \vmu^{t}_j||_2\quad,\quad G_i \in \mathcal{G}^t,\quad\text{and}\quad G_j\in\mathcal{G}_{\text{single}}^{t},
\end{equation}
where we calculate the single-direction Chamfer Distance between the deformed Gaussians $\mathcal{G}^t$ and single-state reconstructed Gaussians $\mathcal{G}^t_\text{single}$ as the loss signal. As these single-state Gaussians are only a rough estimate, we only introduce this loss in the first 1K to 5K steps.
Additionally, to regularize the learning of part centers $\vp_k$, we add another regularization loss as:
\begin{equation}
\mathcal{L}_{\text{reg}}=\frac{1}{K}\sum_{k=1}^K||\vp_k-\hat{\vp}_k||_2, \quad \mathrm{where}\quad \hat{\vp}_k=\sum_{i=1}^{N}\frac{\mM_{ik}}{\sum_{i=1}^{N}\mM_{ik}}\vmu_i,
\end{equation}
which enforces that the centers $\vp_k$ should be close to the average spatial position of Gaussians in canonical Gaussians $\mathcal{G}^c$ that belong to part $k$. Above all, our supervision could be summarized as:
\begin{equation}
\mathcal{L} = \mathcal{L}_{\text{render}} + \lambda_{\text{CD}}\mathcal{L}_{\text{CD}} + \lambda_{\text{reg}}\mathcal{L}_{\text{reg}}.
\end{equation}
We provide more implementation and model training details in~\cref{app:imp}.

\section{Experiments}
\label{sec:exp}


\paragraph{Datasets}
We evaluate our method on three datasets:
(1) PARIS, a two-part dataset proposed by~\cite{jiayi2023paris}, which features articulated objects consisting of one static and one movable part. It includes 10 synthetic objects from the PartNet-Mobility dataset~\citep{xiang2020sapien} and 2 real-world objects captured using the MultiScan~\citep{mao2022multiscan} toolset. 
(2) DTA-Multi, a dataset proposed by~\cite{weng2024neural}, containing 2 synthetic multi-part articulated objects from PartNet-Mobility, each with one static part and two movable parts. 
(3) \model-Multi, our newly curated dataset, featuring 5 complex articulated objects from PartNet-Mobility with 3 to 6 movable parts. 

\paragraph{Metrics} Following the evaluation protocols of PARIS~\citep{jiayi2023paris} and DTA~\citep{weng2024neural}, we assess the performance of all methods using both mesh reconstruction and articulation estimation metrics. 
For mesh reconstruction, we compute the bi-directional Chamfer Distance between the reconstructed mesh and the ground truth mesh with 10K uniformly sampled points from each mesh. We report the Chamfer Distance for the whole object (CD-w), the static parts (CD-s), and the movable parts (CD-m).
For articulation estimation, we evaluate the predicted articulation using the angular error (Axis Ang.) and the distance (Axis Pos., revolute joint only) between the predicted and ground-truth joint axes. We also report the part motion error (Part Motion) which measures the rotation geodesic distance error (in degrees) for revolute joints and Euclidean distance error (in meters) for prismatic joints.


\subsection{Results on Simple Articulated Objects}
\label{sec:exp:two-part}
\paragraph{Experimental Setup} We use the PARIS dataset as the benchmark and select Ditto~\citep{hsu2023ditto}, PARIS~\citep{jiayi2023paris}, CSG-reg~\citep{weng2024neural}, 3Dseg-reg~\citep{weng2024neural}, and DTA~\citep{weng2024neural} as baselines for quantitative evaluation. Following the evaluation setting from DTA~\citep{weng2024neural}, we report all metrics with mean $\pm$ std over 10 trials calculated at the high-visibility joint state. We re-train DTA on the same device (NVIDIA RTX 3090) for training time comparison. Additional results on all joint states are provided in~\cref{tab:app:exp_2part}.

\begin{figure}[t!]
 \centering
 \resizebox{\linewidth}{!}{\includegraphics[width=\linewidth]{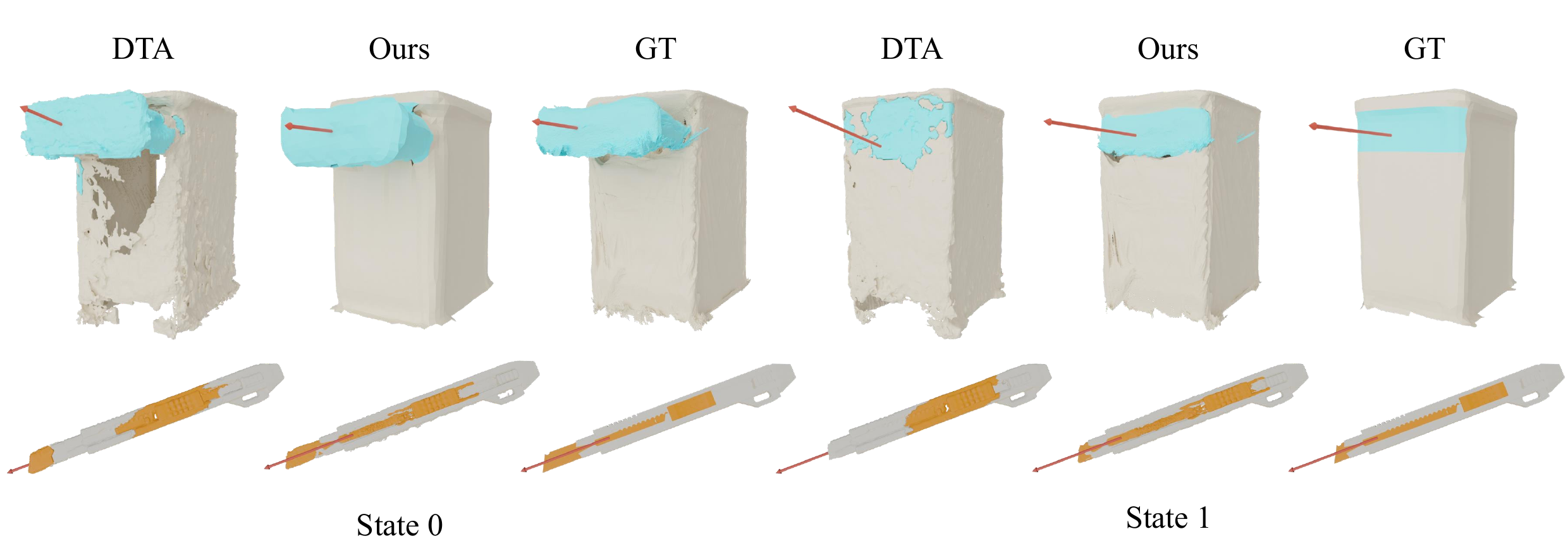}}
 \vspace{-10pt}
 \caption{\textbf{Qualitative visualizations of PARIS objects.} We present reconstruction comparisons between DTA and our model on Real Storage (Top) and Synthetic Blade (Bottom). DTA struggles with mesh reconstruction at the low-visibility state, as it processes each state separately. In contrast, our method leverages the connection between states to improve the reconstruction for both low- and high-visibility states.}
 \label{fig:2part}
 \vspace{-10pt}
\end{figure}

\paragraph{Results}
As shown in \cref{tab:exp_2part}, our method significantly outperforms existing approaches across all metrics, especially for joint articulation parameter estimation, where \model achieves substantially lower errors. This improvement stems from our motion model with Gaussian Splatting, which explicitly deforms Gaussians for more precise part transformation modeling, leading to more precise joint parameter estimation. For mesh reconstruction, \model excels in reconstructing movable parts, yielding lower CD-m values, especially for real-world objects. While DTA performs well on CD-w and CD-s due to its state-by-state reconstruction, we show in~\cref{fig:2part} that it struggles with the low-visibility state. In contrast, \model achieves significantly better results on the low-visibility state while maintaining competitive results on the high-visibility state. This is attributed to the canonical Gaussians modeling that connects the two input joint states for mutually improved mesh reconstruction. Additionally, \model shows consistently better results on real-world objects with significantly faster training time, positioning it as an efficient solution for building digital twins of real-world articulated objects.
\begin{table*}[t!]
\caption{\textbf{Quantitative evaluation on PARIS.} Metrics are reported as mean $\pm$ std over 10 trials at the joint state with higher visibility, following \citep{weng2024neural}. PARIS$^*$~\citep{jiayi2023paris} is augmented with depth for fair comparison. DTA is re-trained for time efficiency comparison. Lower ($\downarrow$) is better on all metrics and we highlight \colorbox[HTML]{ffc5c5}{best} and \colorbox[HTML]{ffebd8}{second best} results. Objects with $\dagger$ are seen categories trained in Ditto. F indicates wrong motion type predictions. Axis Pos. is omitted for prismatic joints (Blade, Storage, and Real Storage).}
\label{tab:exp_2part}
\renewcommand{\arraystretch}{1.2}
\resizebox{\linewidth}{!}{
\begin{tabular}{cc|ccccccccccc|ccc}
\hline
\multirow{2}{*}{Metric} &\multirow{2}{*}{Method} &\multicolumn{11}{c}{Synthetic Objects} &\multicolumn{3}{|c}{Real Objects} \\
& &FoldChair &Fridge &Laptop$^\dagger$ &Oven$^\dagger$ &Scissor &Stapler &USB 
&Washer&Blade &Storage$^\dagger$ &All & Fridge &Storage &All \\
\hline
\multirow{6}{*}{\shortstack{Axis\\Ang}} 
&Ditto
&89.35 &89.30 &3.12 &0.96 &4.50 &89.86 &89.77 &89.51 &79.54 &6.32 &54.22 &\best{1.71} &\secbest{5.88} &\secbest{3.80} \\
&PARIS*
&15.79\tiny{$\pm$29.3} &2.93\tiny{$\pm$5.3} &\secbest{0.03\tiny{$\pm$0.0}} &7.43\tiny{$\pm$23.4} &16.62\tiny{$\pm$32.1} &8.17\tiny{$\pm$15.3} &0.71\tiny{$\pm$0.8} &18.40\tiny{$\pm$23.3} &41.28\tiny{$\pm$31.4} &\secbest{0.03\tiny{$\pm$0.0}} &11.14\tiny{$\pm$16.1} &\secbest{1.90\tiny{$\pm$0.0}} &30.10\tiny{$\pm$10.4} &16.00\tiny{$\pm$5.2} \\
&CSG-reg
&0.10\tiny{$\pm$0.0} &0.27\tiny{$\pm$0.0} &0.47\tiny{$\pm$0.0} &0.35\tiny{$\pm$0.1} &0.28\tiny{$\pm$0.0} &0.30\tiny{$\pm$0.0} &11.78\tiny{$\pm$10.5} &71.93\tiny{$\pm$6.3} &7.64\tiny{$\pm$5.0} &2.82\tiny{$\pm$2.5} &9.60\tiny{$\pm$2.4} &8.92\tiny{$\pm$0.9} &69.71\tiny{$\pm$9.6} &39.31\tiny{$\pm$5.2} \\
&3Dseg-reg
&- &- &2.34\tiny{$\pm$0.11} &- &- &- &- &- &9.40\tiny{$\pm$7.5} &- &- &- &- &- \\
&DTA
&\secbest{0.03\tiny{$\pm$0.0}} &\secbest{0.09\tiny{$\pm$0.0}} &{0.07\tiny{$\pm$0.0}} &\secbest{0.22\tiny{$\pm$0.1}} &\secbest{0.10\tiny{$\pm$0.0}} &\secbest{0.07\tiny{$\pm$0.0}} &\secbest{0.11\tiny{$\pm$0.0}} &\secbest{0.36\tiny{$\pm$0.1}} &\secbest{0.20\tiny{$\pm$0.1}} &{0.09\tiny{$\pm$0.0}} &\secbest{0.13\tiny{$\pm$0.0}} &{2.08\tiny{$\pm$0.0}} 
&13.64\tiny{$\pm$3.6} &7.86\tiny{$\pm$1.8} \\
&Ours
&\best{0.01\tiny{$\pm$0.0}} &\best{0.03\tiny{$\pm$0.0}} &\best{0.01\tiny{$\pm$0.0}} &\best{0.01\tiny{$\pm$0.0}} &\best{0.05\tiny{$\pm$0.0}} &\best{0.01\tiny{$\pm$0.0}} &\best{0.04\tiny{$\pm$0.0}} &\best{0.02\tiny{$\pm$0.0}} &\best{0.03\tiny{$\pm$0.0}} &\best{0.01\tiny{$\pm$0.0}} &\best{0.02\tiny{$\pm$0.0}} & 2.09\tiny{$\pm$0.0} &\best{3.47\tiny{$\pm$0.3}} &\best{2.78\tiny{$\pm$0.2}} \\
\hline
\multirow{6}{*}{\shortstack{Axis\\Pos}} 
&Ditto
&3.77 &1.02 &\secbest{0.01} &0.13 &5.70 &0.20 &5.41 &0.66 &- &- &2.11 &1.84 &- &1.84 \\
&PARIS*
&0.25\tiny{$\pm$0.5} &1.13\tiny{$\pm$2.6} &\best{0.00\tiny{$\pm$0.0}} &0.05\tiny{$\pm$0.2} &1.59\tiny{$\pm$1.7} &4.67\tiny{$\pm$3.9} &3.35\tiny{$\pm$3.1} &3.28\tiny{$\pm$3.1} &- &- &1.79\tiny{$\pm$1.5} &\secbest{0.50\tiny{$\pm$0.0}} &{-} &\secbest{0.50\tiny{$\pm$0.0}} \\
&CSG-reg 
&0.02\tiny{$\pm$0.0} &\best{0.00\tiny{$\pm$0.0}} &0.20\tiny{$\pm$0.2} &0.18\tiny{$\pm$0.0} &\secbest{0.01\tiny{$\pm$0.0}} &\secbest{0.02\tiny{$\pm$0.0}} &\secbest{0.01\tiny{$\pm$0.0}} &2.13\tiny{$\pm$1.5} &- &- &0.32\tiny{$\pm$0.2} &1.46\tiny{$\pm$1.1} &- &1.46\tiny{$\pm$1.1} \\
&3Dseg-reg 
&- &- &0.10\tiny{$\pm$0.0} &- &- &- &- &- &- &- &- &- &- &- \\
&DTA
&\secbest{0.01\tiny{$\pm$0.0}} &\secbest{0.01\tiny{$\pm$0.0}} &\secbest{0.01\tiny{$\pm$0.0}}
&\secbest{0.01\tiny{$\pm$0.0}} &{0.02\tiny{$\pm$0.0}} &\secbest{0.02\tiny{$\pm$0.0}} &\best{0.00\tiny{$\pm$0.0}} &\secbest{0.05\tiny{$\pm$0.0}} 
&{-} &{-} &\secbest{0.02\tiny{$\pm$}0.0} &0.59\tiny{$\pm$0.0}
&- &0.59\tiny{$\pm$0.0} \\
&Ours
&\best{0.00\tiny{$\pm$0.0}} &\best{0.00\tiny{$\pm$0.0}} &\secbest{0.01\tiny{$\pm$0.0}} &\best{0.00\tiny{$\pm$0.0}} &\best{0.00\tiny{$\pm$0.0}} &\best{0.01\tiny{$\pm$0.0}} &\best{0.00\tiny{$\pm$0.0}} &\best{0.00\tiny{$\pm$0.0}} & {-} & {-} &\best{0.00\tiny{$\pm$0.0}} &\best{0.47\tiny{$\pm$0.0}} & {-} &\best{0.47\tiny{$\pm$0.0}} \\
\hline
\multirow{6}{*}{\shortstack{Part\\Motion}}
&Ditto
&99.36 &F &5.18 &2.09 &19.28 &56.61 &80.60 &55.72 &F &0.09 &39.87 &8.43 &0.38 &4.41 \\
&PARIS*
&127.34\tiny{$\pm$75.0} &45.26\tiny{$\pm$58.5} &\secbest{0.03\tiny{$\pm$0.0}} &9.13\tiny{$\pm$28.8} &68.36\tiny{$\pm$64.8} &107.76\tiny{$\pm$68.1} &96.93\tiny{$\pm$67.8} &49.77\tiny{$\pm$26.5} &0.36\tiny{$\pm$0.2} &0.30\tiny{$\pm$0.0} &50.52\tiny{$\pm$39.0} &\best{1.58\tiny{$\pm$0.0}} &0.57\tiny{$\pm$0.1} &1.07\tiny{$\pm$0.1} \\
&CSG-reg 
&0.13\tiny{$\pm$0.0} &0.29\tiny{$\pm$0.0} &0.35\tiny{$\pm$0.0} &0.58\tiny{$\pm$0.0} &\secbest{0.20\tiny{$\pm$0.0}} &0.44\tiny{$\pm$0.0} &10.48\tiny{$\pm$9.3} &158.99\tiny{$\pm$8.8} &\secbest{0.05\tiny{$\pm$0.0}} &\secbest{0.04\tiny{$\pm$0.0}} &17.16\tiny{$\pm$1.8} &14.82\tiny{$\pm$0.1}
&0.64\tiny{$\pm$0.1} &7.73\tiny{$\pm$0.1} \\
&3Dseg-reg
&- &- &1.61\tiny{$\pm$0.1} &- &- &- &- &- &0.15\tiny{$\pm$0.0} &- &- &- &- &- \\
&DTA
&\secbest{0.10\tiny{$\pm$0.0}} &\secbest{0.12\tiny{$\pm$0.0}} &0.11\tiny{$\pm$0.0} &\secbest{0.12\tiny{$\pm$0.0}} &{0.37\tiny{$\pm$0.6}} &\secbest{0.08\tiny{$\pm$0.0}} &\secbest{0.15\tiny{$\pm$0.0}} 
&\secbest{0.28\tiny{$\pm$0.1}} &\best{0.00\tiny{$\pm$0.0}} &\best{0.00\tiny{$\pm$0.0}} &\secbest{0.13\tiny{$\pm$0.1}} &\secbest{1.85\tiny{$\pm$0.0}} &\secbest{0.14\tiny{$\pm$0.0}} &\secbest{1.00\tiny{$\pm$0.0}} \\
&Ours
&\best{0.03\tiny{$\pm$0.0}} &\best{0.04\tiny{$\pm$0.0}}
&\best{0.02\tiny{$\pm$0.0}} &\best{0.02\tiny{$\pm$0.0}}
&\best{0.04\tiny{$\pm$0.0}} &\best{0.01\tiny{$\pm$0.0}}
&\best{0.03\tiny{$\pm$0.0}} &\best{0.03\tiny{$\pm$0.0}}
&\best{0.00\tiny{$\pm$0.0}} &\best{0.00\tiny{$\pm$0.0}}
&\best{0.02\tiny{$\pm$0.0}} & 1.94\tiny{$\pm$0.0}
&\best{0.04\tiny{$\pm$0.0}} &\best{0.99\tiny{$\pm$0.0}} \\
\hline
\multirow{6}{*}{\shortstack{CD-s}} 
&Ditto
&33.79 &3.05 &\secbest{0.25} &\best{2.52} &39.07 &41.64 &2.64 &10.32 &46.90 &9.18 &18.94 &47.01 &16.09 &31.55 \\
&PARIS*
&10.20\tiny{$\pm$5.8} &8.82\tiny{$\pm$12.0} &\best{0.16\tiny{$\pm$0.0}} &\secbest{3.18\tiny{$\pm$0.3}} &15.58\tiny{$\pm$13.3} &\best{2.48\tiny{$\pm$1.9}} &\best{1.95\tiny{$\pm$0.5}} &12.19\tiny{$\pm$3.7} &1.40\tiny{$\pm$0.7} &8.67\tiny{$\pm$0.8} &6.46\tiny{$\pm$3.9} &11.64\tiny{$\pm$1.5} &20.25\tiny{$\pm$2.8} &15.94\tiny{$\pm$2.1} \\
&CSG-reg
&1.69 &1.45 &0.32 &3.93 &\secbest{3.26} &\secbest{2.22} &\best{1.95} &\best{4.53} &0.59 &\secbest{7.06} &2.70 &6.33 &12.55 &9.44 \\
&3Dseg-reg
&- &- &0.76 &- &- &- &- &- &66.31 &- &- &- &- &- \\
&DTA
&\best{0.18\tiny{$\pm$0.0}} &\secbest{0.62\tiny{$\pm$0.0}} &0.30\tiny{$\pm$0.0} &4.60\tiny{$\pm$0.1} &{3.55\tiny{$\pm$6.1}} &{2.91\tiny{$\pm$0.1}} &2.32\tiny{$\pm$0.1} 
&\secbest{4.56\tiny{$\pm$0.1}} &\secbest{0.55\tiny{$\pm$0.0}} &\best{4.90\tiny{$\pm$0.5}} &\best{2.45\tiny{$\pm$0.7}} &\secbest{2.36\tiny{$\pm$0.1}} &\secbest{10.98\tiny{$\pm$0.1}} &\secbest{6.67\tiny{$\pm$0.1}} \\
&Ours
&\secbest{0.26\tiny{$\pm$0.3}} &\best{0.52\tiny{$\pm$0.0}} 
&0.63\tiny{$\pm$0.0} &3.88\tiny{$\pm$0.0} 
&\best{0.61\tiny{$\pm$0.3}} &3.83\tiny{$\pm$0.1} 
&\secbest{2.25\tiny{$\pm$0.2}} &{6.43\tiny{$\pm$0.1}} &\best{0.54\tiny{$\pm$0.0}} &{7.31\tiny{$\pm$0.2}} 
&\secbest{2.63\tiny{$\pm$0.1}} &\best{1.64\tiny{$\pm$0.2}} 
&\best{2.93\tiny{$\pm$0.3}} &\best{2.29\tiny{$\pm$0.3}} \\
\hline
\multirow{6}{*}{\shortstack{CD-m}}
&Ditto
&141.11 &0.99 &\secbest{0.19} &0.94 &20.68 &31.21 &15.88 &12.89 &195.93 &2.20 &42.20 &50.60 &\secbest{20.35} &35.48 \\
&PARIS*
&17.97\tiny{$\pm$24.9} &7.23\tiny{$\pm$11.5} &0.15\tiny{$\pm$0.0} &6.54\tiny{$\pm$10.6} &16.65\tiny{$\pm$16.6} &30.46\tiny{$\pm$37.0} &10.17\tiny{$\pm$6.9} &265.27\tiny{$\pm$248.7} &117.99\tiny{$\pm$213.0} &52.34\tiny{$\pm$11.0} &52.48\tiny{$\pm$58.0} &77.85\tiny{$\pm$26.8} &474.57\tiny{$\pm$227.2} &276.21\tiny{$\pm$127.0} \\
&CSG-reg
&1.91 &21.71 &0.42 &256.99 &\secbest{1.95} &6.36 &29.78 &436.42 &26.62 &1.39 &78.36 &442.17 &521.49 &481.83 \\
&3Dseg-reg
&- &- &1.01 &- &- &- &- &- &6.23 &- &- &- &- &- \\
&DTA
&\best{0.15\tiny{$\pm$0.0}} &\secbest{0.27\tiny{$\pm$0.0}} &\best{0.13\tiny{$\pm$0.0}} &\best{0.44\tiny{$\pm$0.0}} &{10.11\tiny{$\pm$19.4}} &\secbest{1.13\tiny{$\pm$0.5}} &\secbest{1.47\tiny{$\pm$0.0}} 
&\best{0.45\tiny{$\pm$0.0}} &\secbest{2.05\tiny{$\pm$0.3}} &\best{0.36\tiny{$\pm$0.0}} &\secbest{1.66\tiny{$\pm$2.0}} &\secbest{1.12\tiny{$\pm$0.0}} &30.78\tiny{$\pm$2.6} &\secbest{15.95\tiny{$\pm$1.3}} \\
&Ours
&\secbest{0.54\tiny{$\pm$0.1}} &\best{0.21\tiny{$\pm$0.0}} &\best{0.13\tiny{$\pm$0.0}} & \secbest{0.89\tiny{$\pm$0.2}} &\best{0.64\tiny{$\pm$0.4}} &\best{ 0.52\tiny{$\pm$0.1}} &\best{1.22\tiny{$\pm$0.1}} &\best{0.45\tiny{$\pm$0.2}} &\best{1.12\tiny{$\pm$0.2}} & \secbest{1.02\tiny{$\pm$0.4}} &\best{0.67\tiny{$\pm$0.2}} &\best{0.66\tiny{$\pm$0.2}} &\best{6.28\tiny{$\pm$3.6}} &\best{3.47\tiny{$\pm$1.9}} \\

\hline
\multirow{6}{*}{\shortstack{CD-w}}
&Ditto
&6.80 &2.16 &\secbest{0.31} &\best{2.51} &1.70 &2.38 &2.09 &7.29 &42.04 &\best{3.91} &7.12 &6.50 &14.08 &10.29 \\
&PARIS* 
&4.37\tiny{$\pm$6.4} &5.53\tiny{$\pm$4.7} &\best{0.26\tiny{$\pm$0.0}} &{3.18\tiny{$\pm$0.3}} &3.90\tiny{$\pm$3.6} &5.27\tiny{$\pm$5.9} &1.78\tiny{$\pm$0.2} &10.11\tiny{$\pm$2.8} &0.58\tiny{$\pm$0.1} &7.80\tiny{$\pm$0.4} &4.28\tiny{$\pm$2.4} &8.99\tiny{$\pm$1.4} &32.10\tiny{$\pm$8.2} &20.55\tiny{$\pm$4.8} \\
&CSG-reg
&0.48 &0.98 &0.40 &\secbest{3.00} &1.70 &\secbest{1.99} &\secbest{1.20} &\best{4.48} &\secbest{0.56} &4.00 &\secbest{1.88} &5.71 &14.29 &10.00 \\
&3Dseg-reg
&- &- &0.81 &- &- &- &- &- &0.78 &- &- &- &- &- \\
&DTA
&\best{0.27\tiny{$\pm$0.0}} &\secbest{0.70\tiny{$\pm$0.0}} &0.32\tiny{$\pm$0.0} &4.24\tiny{$\pm$0.1} &\best{0.41\tiny{$\pm$0.0}} &\best{1.92\tiny{$\pm$0.0}} &\best{1.17\tiny{$\pm$0.0} }
&\best{4.48\tiny{$\pm$0.2}} &\best{0.36\tiny{$\pm$0.0}} &\secbest{3.99\tiny{$\pm$0.4}} &\best{1.79\tiny{$\pm$0.1}} &\secbest{2.08\tiny{$\pm$0.1}} &\secbest{8.98\tiny{$\pm$0.1}} &\secbest{5.53\tiny{$\pm$0.1}} \\
&Ours
&\secbest{0.43\tiny{$\pm$0.2}} &\best{0.58\tiny{$\pm$0.0}} & 0.50\tiny{$\pm$0.0} &3.58\tiny{$\pm$0.0} & \secbest{0.67\tiny{$\pm$0.3}} & {2.63\tiny{$\pm$0.0}} & {1.28\tiny{$\pm$0.0}} & \secbest{5.99\tiny{$\pm$0.1}} & 0.61\tiny{$\pm$0.0} & 5.21\tiny{$\pm$0.1} & {2.15\tiny{$\pm$0.1}} &\best{1.29\tiny{$\pm$0.1}} &\best{3.23\tiny{$\pm$0.1}} &\best{2.26\tiny{$\pm$0.1}} \\
\hline
\multirow{2}{*}{\shortstack{Time\\(min)}}
&DTA
&29 &30 &31 &29 &28 &29 &31 &28 &27 &28 &29 &29 &29 &29 \\
&Ours 
&9 &8 &7 &7 &7 &7 &7 &8 &7 &8 &8 &9 &9 &9 \\
\hline
\end{tabular}
}
\vspace{-10pt}
\end{table*}



\subsection{Results on Complex Articulated Objects with Multiple Movable Parts}
\label{sec:exp:multi-part}

\paragraph{Experimental Setup} We use DTA-Multi and \model-Multi as benchmarks for evaluating complex articulated object reconstruction. On DTA-Multi, we compare our model against PARIS and DTA, while on \model-Multi we use DTA as the main baseline given its strong performance. Similar to~\cref{sec:exp:two-part}, we report all metrics with a mean over 10 trials for DTA-Multi and 3 trials for \model-multi because of the training time required for baselines. For \model-multi, we report the average of all movable parts for articulation estimation and mesh reconstruction due to the large number of parts. Considering the potential error prediction with no mesh for one of the parts, we manually set the Chamfer Distance of the empty prediction to 1000.

\begin{table*}[t]
\caption{\textbf{Quantitative evaluation on DTA-Multi.} We report averaged metrics over 10 trials with different random seeds. Lower ($\downarrow$) is better on all metrics. Joint 1 of ``Storage-m'' is prismatic with no Axis Pos.}
\label{tab:exp_mpart_dta}
\renewcommand{\arraystretch}{1.2}
\resizebox{\linewidth}{!}{
\begin{tabular}{ccccccccccccc}
\toprule
Object &Method
&Axis Ang 0 &Axis Ang 1 &Axis Pos 0 &Axis Pos 1 
&Part Motion 0 &Part Motion 1 &CD-s &CD-m 0 &CD-m 1 &CD-w &Time (min)\\
\midrule
\multirow{3}{*}{Fridge-m} 
&PARIS
&34.52 &15.91 &3.60 &1.63 &86.21 &105.86 &8.52 &526.19 &160.86 &15.00 &- \\
&DTA
&0.25 &0.06 &0.01 &0.01 &0.23 &0.08 &0.63 &0.44 &0.53 &0.88 &32 \\
&Ours 
&\textbf{0.02} &\textbf{0.00} &\textbf{0.00} &\textbf{0.00} &\textbf{0.02} &\textbf{0.03} & \textbf{0.62} &\textbf{0.07} &\textbf{0.18} &\textbf{0.75} &\textbf{8} \\
\midrule
\multirow{3}{*}{Storage-m} 
&PARIS
&43.26 &26.18 &10.42 &- &79.84 &0.64 &8.56 &128.62 &266.71 &8.66 &- \\
&DTA
&0.17 &0.40 &0.04 &- &0.13 &0.00 &0.86 &0.20 & \textbf{0.25} &0.97 &32 \\
&Ours
&\textbf{0.01} & \textbf{0.02} & \textbf{0.01} & - & \textbf{0.01} & \textbf{0.00} & \textbf{0.78} & \textbf{0.19} & 0.27 & \textbf{0.93} &\textbf{8} \\
 \bottomrule
\end{tabular}
}
\vspace{-5pt}
\end{table*}

\begin{table*}[t]
\caption{\textbf{Quantitative evaluation on \model-Multi}. Metrics are averaged over 3 trials. Due to the large number of parts, we report the average metric for all movable parts. Lower ($\downarrow$) is better on all metrics. ``Table-31249'' has 3 prismatic joints with no Axis Pos. }
\label{tab:exp_mpart_our}
\renewcommand{\arraystretch}{1.2}
\resizebox{\linewidth}{!}{
\begin{tabular}{ccccccccccc}
\toprule
Object &Method
&Axis Ang &Axis Pos &Part Motion &CD-s &CD-m &CD-w &Time (min)\\
\midrule
\multirow{2}{*}{\shortstack{Table \\ \scriptsize{25493 (4 parts)}}}
&DTA 
&24.35 &- &0.12 &\textbf{0.59} &104.38 &\textbf{0.55} &34 \\
&Ours 
&\textbf{1.16} &- &\textbf{0.00} &0.74 &\textbf{3.53} &0.74 &\textbf{8} \\
\midrule
\multirow{2}{*}{\shortstack{Table \\ \scriptsize{31249 (5 parts)}}}
&DTA 
&20.62 &4.2 &30.8 &1.39 &230.38 &\textbf{1.00} &37 \\
&Ours 
&\textbf{0.04} &\textbf{0.00} &\textbf{0.01} &\textbf{1.22} &\textbf{3.09} &1.16 &\textbf{8} \\
\midrule
\multirow{2}{*}{\shortstack{Storage \\ \scriptsize{45503 (4 parts)}}}
&DTA 
&51.18 &2.44 &43.77 &5.74 &246.63 &0.88 &35 \\
&Ours 
&\textbf{0.02} &\textbf{0.00} &\textbf{0.03} &\textbf{0.75} &\textbf{0.13} &\textbf{0.88} &\textbf{8} \\
\midrule
\multirow{2}{*}{\shortstack{Storage \\ \scriptsize{47468 (7 parts)}}}
&DTA
&19.07 &0.31 &10.67 &0.82 &476.91 &0.71 &45 \\
&Ours 
&\textbf{0.14} &\textbf{0.02} &\textbf{0.62} &\textbf{0.67} &\textbf{3.70} &\textbf{0.70} &\textbf{8} \\
\midrule
\multirow{2}{*}{\shortstack{Oven \\ \scriptsize{101908 (4 parts)}}}
&DTA 
&17.83 &6.51 &31.80 &1.17 &359.16 &\textbf{1.01} &35 \\
&Ours 
&\textbf{0.04} &\textbf{0.01} &\textbf{0.23} &\textbf{1.08} &\textbf{0.25} &1.03 &\textbf{8} \\
\bottomrule 
\end{tabular}
}
\vspace{-5pt}
\end{table*}
\begin{figure}[t!]
 \centering
 \resizebox{\linewidth}{!}{\includegraphics[width=\linewidth]{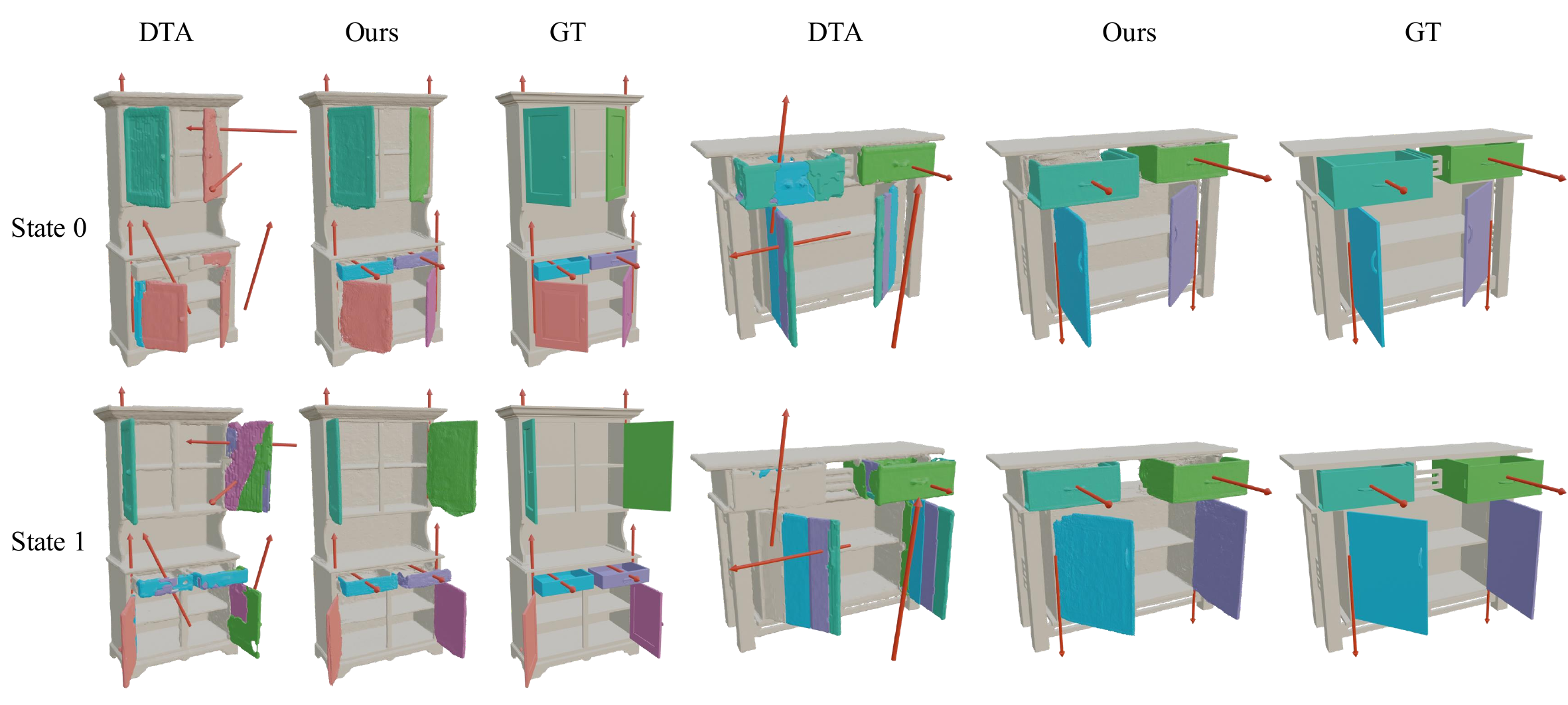}}
 \caption{\textbf{Qualitative results on multi-part objects.} We present reconstruction comparisons between DTA and our model on Storage-47648 (Left) and Table-31249 (Bottom). On \model-Multi, DTA struggles with movable part identification and axis prediction as the number of parts increases, whereas our model maintains high performance regardless of part count, achieving high-quality reconstruction of part mesh and joint articulation.}
 \label{fig:mpart}
 \vspace{-10pt}
\end{figure}

\paragraph{Results} As demonstrated in~\cref{tab:exp_mpart_dta} and~\cref{tab:exp_mpart_our}, our method consistently outperforms existing methods by a large margin in both mesh reconstruction and articulation estimation. Notably, on \model-Multi, the baseline model DTA struggles with movable part identification and axis prediction as the number of parts increases, whereas our model maintains high performance regardless of part count. We also provide a qualitative comparison in~\cref{fig:mpart} for better visualization. Moreover, our method maintains the same time efficiency while the training time of existing methods scales with the number of parts. These results underscore the robustness and effectiveness of our method in modeling complex, multi-part articulated objects.





\subsection{Ablative Studies}
\label{sec:exp:ablation}
\paragraph{Experimental Setup} To verify the effectiveness of our model design, we meticulously design four ablations of \model to identify the impact of key components in our method: (i) Randomly initializing canonical Gaussians (\textit{w/o} Cano. Init.), (ii) predicting part assignments with MLP (\textit{w/} MLP Seg) or Slot-Attention (\textit{w/} SA Seg), (iii) randomly initializing part centers $C_k$ (\textit{w/o} Center Init.), (iv) clustering all Gaussians instead of clustering movable Gaussians for part center initialization (\textit{w/o} Motion Prior), and (v) learning articulation parameters without the joint prediction warmup stage (\textit{w/o} Joint Pred.). We select two representative objects: ``Storage-47648'' with 4 revolute and 2 prismatic joints and ``Oven-101908'' with 3 revolute joints for ablative analysis. Similar to~\cref{sec:exp:multi-part}, we report the average of all parts over 10 trials for all metrics.

\begin{table}[t]
\centering
\caption{\textbf{Ablative experiments}. Lower ($\downarrow$) is better on all metrics.}
\label{tab:ablation}
\renewcommand{\arraystretch}{1.2}
\resizebox{\linewidth}{!}{
\begin{tabular}{lcccccccccccc}
\toprule
\multirow{2}[2]{*}{Method} & \multicolumn{6}{c}{Storage 47648 (7 parts)} & \multicolumn{6}{c}{Oven 101908 (4 parts)} \\
\cmidrule(lr){2-7}\cmidrule{8-13}
&Axis Ang &Axis Pos &Part Motion &CD-s &CD-m & CD-w & Axis Ang &Axis Pos &Part Motion &CD-s &CD-m &CD-w \\
\midrule
Full
&\textbf{0.14} &\textbf{0.02} &\textbf{0.62} &\textbf{0.67} &\textbf{3.70} &\textbf{0.70} 
&\textbf{0.04} &\textbf{0.01} &\textbf{0.23} &\textbf{1.08} &\textbf{0.25} &\textbf{1.03} \\
\textit{w/o} Cano. init.
&24.15 & 0.73 & 20.61 & 0.83 & 495.07 & 1.25 
& 57.87 & 2.95 & 54.45 & 1.73 & 1030.19 & 2.36 \\
\textit{w/o} Center Init.
& 52.78 & 0.83 & 33.04 & 1.09 & 344.19 & 1.69 
& 28.94 & 2.36 & 22.46 & 1.41 & 8.86 & 2.13 \\ 
\textit{w/o} Motion Prior
& 26.74 & 0.22 & 21.16 & 258.23 & 599.46 & 1.15 
& 40.08 & 0.98 & 41.06 & 1.75 & 503.44 & 2.35 \\
\textit{w/o} Joint Pred. 
&{0.16} &{0.02} &{0.72} &{0.67} &{3.90} &{0.71}
&{0.04} &{0.01} &{0.23} &{1.08} &{0.25} &{1.03} \\
\textit{w/} MLP Seg 
& 21.84 & 3.46 & 31.43 & 1.82 & 664.25 & 1.28 
& 12.08 & 3.33 & 27.28 & 7.78 & 126.95 & 2.19 \\
\textit{w/} SA Seg 
& 25.43 & 0.7 & 23.22 & 1.52 & 459.89 & 1.16 
& 58.04 & 4.53 & 51.28 & 1.26 & 496.64 & 2.35 \\
\bottomrule
\end{tabular}
}
\vspace{-10pt}
\end{table}

\begin{figure}[t!]
 \centering
 \resizebox{\linewidth}{!}{\includegraphics[width=\linewidth]{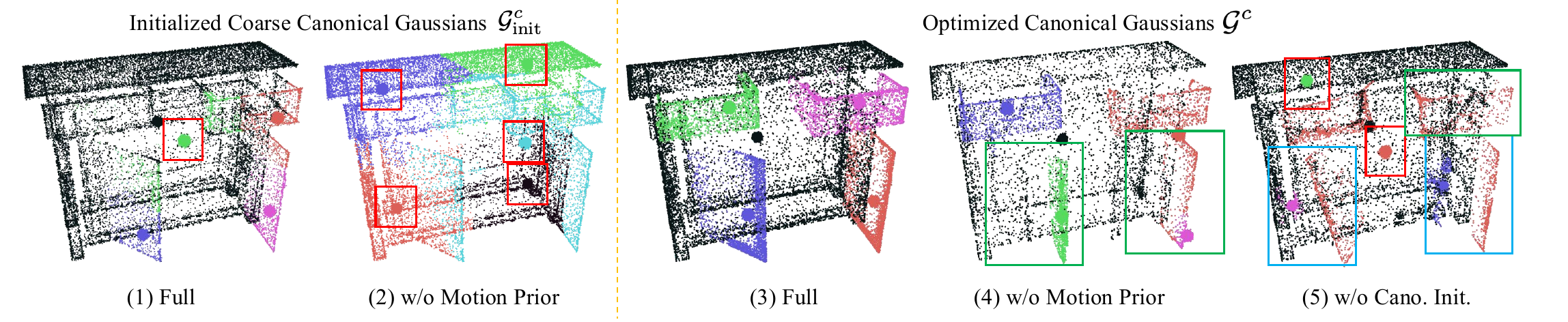}}
 \caption{\textbf{Abaltion Studies}. We visualize the initialized and optimized canonical Gaussians with their part assignment and centers for the full model, w/o Motion Prior and w/o Cano. Init. We highlight center error, part assignment error, and canonical Gaussian error with \textcolor{red}{red}, \textcolor{green}{green}, and \textcolor{blue}{blue} bounding boxes separately.}
 \label{fig:ablation}
 \vspace{-10pt}
\end{figure}

\paragraph{Results and Discussions} As shown in~\cref{tab:ablation} and \cref{fig:ablation}, we make the following observations:
\begin{itemize}[leftmargin=*,nolistsep,noitemsep]
\item\textit{Canonical Gaussians Initialization.} Omitting this initialization strategy significantly degrades the model performance across all metrics, particularly for movable parts. As illustrated in \cref{fig:ablation} (5), the absence of our initialization strategy leads to malformed canonical Gaussians, making the model converge to suboptimal local minima during optimization.

\item\textit{Center-based Part Modeling and Assignment.} Replacing our center-based part assignment module with MLP or Slot-Attention ("w/ MLP Seg" and "w/ SA Seg") leads to substantial performance drops, especially in joint parameter estimation and movable part reconstruction. This demonstrates the superiority of our center-based approach in accurately segmenting articulated parts.

\item\textit{Center Initialization.} Random center initialization performs well for static parts but poorly for movable parts. Clustering all Gaussians fails to reconstruct both static and movable parts due to incorrect center initialization. As illustrated in \cref{fig:ablation} (1), clustering on movable Gaussians still produces an incorrect center but provides a good starting point for optimization. Our \model will refine the centers in the optimization process as shown in \cref{fig:ablation} (3). In contrast, clustering on all Gaussians results in entirely wrong center initialization (\cref{fig:ablation} (2)), which is difficult to correct (\cref{fig:ablation} (4)), leading to even worse performance than random initialization. This highlights the importance of our center initialization strategy in achieving accurate part articulation modeling.

\item\textit{Joint Prediction Warmup.} This technique primarily affects prismatic joints, as we do not constrain the transformation of revolute joints. As shown in \cref{tab:ablation}, predicting the joint type and then refining joint parameters with type constraints slightly improves the articulation reconstruction.
\end{itemize}

In summary, these ablation studies confirm that each component contributes significantly to its overall performance, playing crucial roles in achieving accurate joint parameter estimation and high-quality part mesh reconstruction. We provide further discussions in~\cref{app:discussion} and ~\cref{app:limitation}.

\section{Conclusion}
\label{sec:conclusion}
In conclusion, we propose \model, a novel approach for reconstructing articulated objects from two states of multi-view images. By leveraging 3D Gaussians and introducing novel techniques for state alignment and part dynamics modeling, our approach overcomes key limitations of existing methods. The performance improvements in joint parameter estimation and part mesh reconstruction, particularly for complex multi-part objects, demonstrate the effectiveness of our innovations. Our comprehensive experiments across synthetic and real-world datasets validate the robustness and efficiency of \model, while also revealing promising directions for future research. As the demand for accurate digital replicas of articulated objects continues to grow in fields such as robotics and augmented reality, \model provides a solid foundation for bridging the gap between physical and virtual environments. Moving forward, we anticipate that the principles introduced in this work will inspire further advancements in the field, ultimately enabling more sophisticated and realistic simulations for a wide range of applications.


\begin{thebibliography}{103}
\providecommand{\natexlab}[1]{#1}
\providecommand{\url}[1]{\texttt{#1}}
\expandafter\ifx\csname urlstyle\endcsname\relax
  \providecommand{\doi}[1]{doi: #1}\else
  \providecommand{\doi}{doi: \begingroup \urlstyle{rm}\Url}\fi

\bibitem[Bae et~al.(2024)Bae, Kim, Yun, Lee, Bang, and Uh]{bae2024per}
Jeongmin Bae, Seoha Kim, Youngsik Yun, Hahyun Lee, Gun Bang, and Youngjung Uh.
\newblock Per-gaussian embedding-based deformation for deformable 3d gaussian splatting.
\newblock \emph{arXiv preprint arXiv:2404.03613}, 2024.

\bibitem[Chen et~al.(2024{\natexlab{a}})Chen, Li, Ye, Wang, Xie, Zhai, Wang, Liu, Bao, and Zhang]{chen2024pgsr}
Danpeng Chen, Hai Li, Weicai Ye, Yifan Wang, Weijian Xie, Shangjin Zhai, Nan Wang, Haomin Liu, Hujun Bao, and Guofeng Zhang.
\newblock Pgsr: Planar-based gaussian splatting for efficient and high-fidelity surface reconstruction.
\newblock \emph{arXiv preprint arXiv:2406.06521}, 2024{\natexlab{a}}.

\bibitem[Chen et~al.(2024{\natexlab{b}})Chen, Ni, Jiang, Zhang, Zhu, and Huang]{chen2024single}
Yixin Chen, Junfeng Ni, Nan Jiang, Yaowei Zhang, Yixin Zhu, and Siyuan Huang.
\newblock Single-view 3d scene reconstruction with high-fidelity shape and texture.
\newblock In \emph{Proceedings of International Conference on 3D Vision (3DV)}, 2024{\natexlab{b}}.

\bibitem[Chu et~al.(2023)Chu, Liu, Ye, Tan, Qi, Fu, and Jia]{chu2023command}
Ruihang Chu, Zhengzhe Liu, Xiaoqing Ye, Xiao Tan, Xiaojuan Qi, Chi-Wing Fu, and Jiaya Jia.
\newblock Command-driven articulated object understanding and manipulation.
\newblock In \emph{Proceedings of Conference on Computer Vision and Pattern Recognition (CVPR)}, 2023.

\bibitem[Deng et~al.(2024)Deng, Subr, and Bilen]{deng2024articulate}
Jianning Deng, Kartic Subr, and Hakan Bilen.
\newblock Articulate your nerf: Unsupervised articulated object modeling via conditional view synthesis.
\newblock \emph{arXiv preprint arXiv:2406.16623}, 2024.

\bibitem[Gadre et~al.(2021)Gadre, Ehsani, and Song]{gadre2021act}
Samir~Yitzhak Gadre, Kiana Ehsani, and Shuran Song.
\newblock Act the part: Learning interaction strategies for articulated object part discovery.
\newblock In \emph{Proceedings of International Conference on Computer Vision (ICCV)}, 2021.

\bibitem[Geng et~al.(2023{\natexlab{a}})Geng, Li, Geng, Chen, Dong, and Wang]{geng2023partmanip}
Haoran Geng, Ziming Li, Yiran Geng, Jiayi Chen, Hao Dong, and He~Wang.
\newblock Partmanip: Learning cross-category generalizable part manipulation policy from point cloud observations.
\newblock In \emph{Proceedings of Conference on Computer Vision and Pattern Recognition (CVPR)}, 2023{\natexlab{a}}.

\bibitem[Geng et~al.(2023{\natexlab{b}})Geng, Xu, Zhao, Xu, Yi, Huang, and Wang]{geng2023gapartnet}
Haoran Geng, Helin Xu, Chengyang Zhao, Chao Xu, Li~Yi, Siyuan Huang, and He~Wang.
\newblock Gapartnet: Cross-category domain-generalizable object perception and manipulation via generalizable and actionable parts.
\newblock In \emph{Proceedings of Conference on Computer Vision and Pattern Recognition (CVPR)}, 2023{\natexlab{b}}.

\bibitem[Gong et~al.(2023)Gong, Huang, Zhao, Geng, Gao, Wu, Ai, Zhou, Terzopoulos, Zhu, et~al.]{gong2023arnold}
Ran Gong, Jiangyong Huang, Yizhou Zhao, Haoran Geng, Xiaofeng Gao, Qingyang Wu, Wensi Ai, Ziheng Zhou, Demetri Terzopoulos, Song-Chun Zhu, et~al.
\newblock Arnold: A benchmark for language-grounded task learning with continuous states in realistic 3d scenes.
\newblock In \emph{Proceedings of International Conference on Computer Vision (ICCV)}, 2023.

\bibitem[Guo et~al.(2024)Guo, Zhou, Li, Wang, and Li]{guo2024motion}
Zhiyang Guo, Wengang Zhou, Li~Li, Min Wang, and Houqiang Li.
\newblock Motion-aware 3d gaussian splatting for efficient dynamic scene reconstruction.
\newblock \emph{arXiv preprint arXiv:2403.11447}, 2024.

\bibitem[Heppert et~al.(2023)Heppert, Irshad, Zakharov, Liu, Ambrus, Bohg, Valada, and Kollar]{heppert2023carto}
Nick Heppert, Muhammad~Zubair Irshad, Sergey Zakharov, Katherine Liu, Rares~Andrei Ambrus, Jeannette Bohg, Abhinav Valada, and Thomas Kollar.
\newblock Carto: Category and joint agnostic reconstruction of articulated objects.
\newblock In \emph{Proceedings of Conference on Computer Vision and Pattern Recognition (CVPR)}, 2023.

\bibitem[Hsu et~al.(2023)Hsu, Jiang, and Zhu]{hsu2023ditto}
Cheng-Chun Hsu, Zhenyu Jiang, and Yuke Zhu.
\newblock Ditto in the house: Building articulation models of indoor scenes through interactive perception.
\newblock In \emph{Proceedings of International Conference on Robotics and Automation (ICRA)}, 2023.

\bibitem[Hu et~al.(2017)Hu, Li, Van~Kaick, Shamir, Zhang, and Huang]{hu2017learning}
Ruizhen Hu, Wenchao Li, Oliver Van~Kaick, Ariel Shamir, Hao Zhang, and Hui Huang.
\newblock Learning to predict part mobility from a single static snapshot.
\newblock \emph{ACM Transactions on Graphics (TOG)}, 36\penalty0 (6):\penalty0 1--13, 2017.

\bibitem[Huang et~al.(2024{\natexlab{a}})Huang, Yu, Chen, Geiger, and Gao]{huang20242d}
Binbin Huang, Zehao Yu, Anpei Chen, Andreas Geiger, and Shenghua Gao.
\newblock 2d gaussian splatting for geometrically accurate radiance fields.
\newblock In \emph{ACM SIGGRAPH 2024 Conference Papers}, 2024{\natexlab{a}}.

\bibitem[Huang et~al.(2024{\natexlab{b}})Huang, Yong, Ma, Linghu, Li, Wang, Li, Zhu, Jia, and Huang]{huang2024embodied}
Jiangyong Huang, Silong Yong, Xiaojian Ma, Xiongkun Linghu, Puhao Li, Yan Wang, Qing Li, Song-Chun Zhu, Baoxiong Jia, and Siyuan Huang.
\newblock An embodied generalist agent in 3d world.
\newblock In \emph{Proceedings of International Conference on Machine Learning (ICML)}, 2024{\natexlab{b}}.

\bibitem[Huang et~al.(2014)Huang, Walker, and Birchfield]{huang2014occlusion}
Xiaoxia Huang, Ian Walker, and Stan Birchfield.
\newblock Occlusion-aware multi-view reconstruction of articulated objects for manipulation.
\newblock \emph{Robotics and Autonomous Systems}, 62\penalty0 (4):\penalty0 497--505, 2014.

\bibitem[Huang et~al.(2024{\natexlab{c}})Huang, Sun, Yang, Lyu, Cao, and Qi]{huang2024sc}
Yi-Hua Huang, Yang-Tian Sun, Ziyi Yang, Xiaoyang Lyu, Yan-Pei Cao, and Xiaojuan Qi.
\newblock Sc-gs: Sparse-controlled gaussian splatting for editable dynamic scenes.
\newblock In \emph{Proceedings of Conference on Computer Vision and Pattern Recognition (CVPR)}, 2024{\natexlab{c}}.

\bibitem[Jain et~al.(2021)Jain, Lioutikov, Chuck, and Niekum]{jain2021screwnet}
Ajinkya Jain, Rudolf Lioutikov, Caleb Chuck, and Scott Niekum.
\newblock Screwnet: Category-independent articulation model estimation from depth images using screw theory.
\newblock In \emph{Proceedings of International Conference on Robotics and Automation (ICRA)}, 2021.

\bibitem[Jia et~al.(2023)Jia, Liu, and Huang]{jia2023improving}
Baoxiong Jia, Yu~Liu, and Siyuan Huang.
\newblock Improving object-centric learning with query optimization.
\newblock In \emph{Proceedings of International Conference on Learning Representations (ICLR)}, 2023.

\bibitem[Jia et~al.(2024)Jia, Chen, Yu, Wang, Niu, Liu, Li, and Huang]{jia2024sceneverse}
Baoxiong Jia, Yixin Chen, Huangyue Yu, Yan Wang, Xuesong Niu, Tengyu Liu, Qing Li, and Siyuan Huang.
\newblock Sceneverse: Scaling 3d vision-language learning for grounded scene understanding.
\newblock In \emph{Proceedings of European Conference on Computer Vision (ECCV)}, 2024.

\bibitem[Jiang et~al.(2024)Jiang, Yu, Xie, Li, Feng, Wang, Li, Lau, Gao, Yang, et~al.]{jiang2024vr}
Ying Jiang, Chang Yu, Tianyi Xie, Xuan Li, Yutao Feng, Huamin Wang, Minchen Li, Henry Lau, Feng Gao, Yin Yang, et~al.
\newblock Vr-gs: a physical dynamics-aware interactive gaussian splatting system in virtual reality.
\newblock In \emph{ACM SIGGRAPH 2024 Conference Papers}, 2024.

\bibitem[Jiang et~al.(2022)Jiang, Hsu, and Zhu]{jiang2022ditto}
Zhenyu Jiang, Cheng-Chun Hsu, and Yuke Zhu.
\newblock Ditto: Building digital twins of articulated objects from interaction.
\newblock In \emph{Proceedings of Conference on Computer Vision and Pattern Recognition (CVPR)}, 2022.

\bibitem[Joo et~al.(2018)Joo, Simon, and Sheikh]{joo2018total}
Hanbyul Joo, Tomas Simon, and Yaser Sheikh.
\newblock Total capture: A 3d deformation model for tracking faces, hands, and bodies.
\newblock In \emph{Proceedings of Conference on Computer Vision and Pattern Recognition (CVPR)}, 2018.

\bibitem[Jung et~al.(2023)Jung, Brasch, Song, Perez-Pellitero, Zhou, Li, Navab, and Busam]{jung2023deformable}
HyunJun Jung, Nikolas Brasch, Jifei Song, Eduardo Perez-Pellitero, Yiren Zhou, Zhihao Li, Nassir Navab, and Benjamin Busam.
\newblock Deformable 3d gaussian splatting for animatable human avatars.
\newblock \emph{arXiv preprint arXiv:2312.15059}, 2023.

\bibitem[Katsumata et~al.(2023)Katsumata, Vo, and Nakayama]{katsumata2023efficient}
Kai Katsumata, Duc~Minh Vo, and Hideki Nakayama.
\newblock An efficient 3d gaussian representation for monocular/multi-view dynamic scenes.
\newblock \emph{arXiv preprint arXiv:2311.12897}, 2023.

\bibitem[Katz et~al.(2013)Katz, Kazemi, Bagnell, and Stentz]{katz2013interactive}
Dov Katz, Moslem Kazemi, J~Andrew Bagnell, and Anthony Stentz.
\newblock Interactive segmentation, tracking, and kinematic modeling of unknown 3d articulated objects.
\newblock In \emph{Proceedings of International Conference on Robotics and Automation (ICRA)}, 2013.

\bibitem[Kawana et~al.(2021)Kawana, Mukuta, and Harada]{kawana2021unsupervised}
Yuki Kawana, Yusuke Mukuta, and Tatsuya Harada.
\newblock Unsupervised pose-aware part decomposition for 3d articulated objects.
\newblock \emph{arXiv preprint arXiv:2110.04411}, 2021.

\bibitem[Kerbl et~al.(2023)Kerbl, Kopanas, Leimk{\"u}hler, and Drettakis]{kerbl20233d}
Bernhard Kerbl, Georgios Kopanas, Thomas Leimk{\"u}hler, and George Drettakis.
\newblock 3d gaussian splatting for real-time radiance field rendering.
\newblock \emph{ACM Trans. Graph.}, 42\penalty0 (4):\penalty0 139--1, 2023.

\bibitem[Kerr et~al.(2024)Kerr, Kim, Wu, Yi, Wang, Goldberg, and Kanazawa]{kerr2024rsrd}
Justin Kerr, Chung~Min Kim, Mingxuan Wu, Brent Yi, Qianqian Wang, Ken Goldberg, and Angjoo Kanazawa.
\newblock Robot see robot do: Imitating articulated object manipulation with monocular 4d reconstruction.
\newblock In \emph{Conference on Robot Learning (CoRL)}, 2024.

\bibitem[Kirillov et~al.(2023)Kirillov, Mintun, Ravi, Mao, Rolland, Gustafson, Xiao, Whitehead, Berg, Lo, et~al.]{kirillov2023segment}
Alexander Kirillov, Eric Mintun, Nikhila Ravi, Hanzi Mao, Chloe Rolland, Laura Gustafson, Tete Xiao, Spencer Whitehead, Alexander~C Berg, Wan-Yen Lo, et~al.
\newblock Segment anything.
\newblock In \emph{Proceedings of International Conference on Computer Vision (ICCV)}, 2023.

\bibitem[Lei et~al.(2023)Lei, Deng, Shen, Guibas, and Daniilidis]{lei2023nap}
Jiahui Lei, Congyue Deng, William~B Shen, Leonidas~J Guibas, and Kostas Daniilidis.
\newblock Nap: Neural 3d articulated object prior.
\newblock 2023.

\bibitem[Lei et~al.(2024)Lei, Wang, Pavlakos, Liu, and Daniilidis]{lei2024gart}
Jiahui Lei, Yufu Wang, Georgios Pavlakos, Lingjie Liu, and Kostas Daniilidis.
\newblock Gart: Gaussian articulated template models.
\newblock In \emph{Proceedings of Conference on Computer Vision and Pattern Recognition (CVPR)}, 2024.

\bibitem[Lewis et~al.(2022)Lewis, Pavlasek, and Jenkins]{lewis2022narf22}
Stanley Lewis, Jana Pavlasek, and Odest~Chadwicke Jenkins.
\newblock Narf22: Neural articulated radiance fields for configuration-aware rendering.
\newblock In \emph{Proceedings of International Conference on Intelligent Robots and Systems (IROS)}, 2022.

\bibitem[Li et~al.(2020)Li, Wang, Yi, Guibas, Abbott, and Song]{li2020category}
Xiaolong Li, He~Wang, Li~Yi, Leonidas~J Guibas, A~Lynn Abbott, and Shuran Song.
\newblock Category-level articulated object pose estimation.
\newblock In \emph{Proceedings of Conference on Computer Vision and Pattern Recognition (CVPR)}, 2020.

\bibitem[Li et~al.(2024)Li, Chen, Li, and Xu]{li2024spacetime}
Zhan Li, Zhang Chen, Zhong Li, and Yi~Xu.
\newblock Spacetime gaussian feature splatting for real-time dynamic view synthesis.
\newblock In \emph{Proceedings of Conference on Computer Vision and Pattern Recognition (CVPR)}, 2024.

\bibitem[Linghu et~al.(2024)Linghu, Huang, Niu, Ma, Jia, and Huang]{linghu2024multi}
Xiongkun Linghu, Jiangyong Huang, Xuesong Niu, Xiaojian Ma, Baoxiong Jia, and Siyuan Huang.
\newblock Multi-modal situated reasoning in 3d scenes.
\newblock \emph{Proceedings of Advances in Neural Information Processing Systems (NeurIPS)}, 2024.

\bibitem[Liu et~al.(2023{\natexlab{a}})Liu, Mahdavi-Amiri, and Savva]{jiayi2023paris}
Jiayi Liu, Ali Mahdavi-Amiri, and Manolis Savva.
\newblock Paris: Part-level reconstruction and motion analysis for articulated objects.
\newblock In \emph{Proceedings of International Conference on Computer Vision (ICCV)}, 2023{\natexlab{a}}.

\bibitem[Liu et~al.(2024)Liu, Tam, Mahdavi-Amiri, and Savva]{liu2024cage}
Jiayi Liu, Hou In~Ivan Tam, Ali Mahdavi-Amiri, and Manolis Savva.
\newblock Cage: Controllable articulation generation.
\newblock In \emph{Proceedings of Conference on Computer Vision and Pattern Recognition (CVPR)}, 2024.

\bibitem[Liu et~al.(2022{\natexlab{a}})Liu, Xu, Fu, Qian, Yu, Han, and Lu]{liu2022akb}
Liu Liu, Wenqiang Xu, Haoyuan Fu, Sucheng Qian, Qiaojun Yu, Yang Han, and Cewu Lu.
\newblock Akb-48: A real-world articulated object knowledge base.
\newblock In \emph{Proceedings of Conference on Computer Vision and Pattern Recognition (CVPR)}, 2022{\natexlab{a}}.

\bibitem[Liu et~al.(2022{\natexlab{b}})Liu, Xue, Xu, Fu, and Lu]{liu2022toward}
Liu Liu, Han Xue, Wenqiang Xu, Haoyuan Fu, and Cewu Lu.
\newblock Toward real-world category-level articulation pose estimation.
\newblock \emph{Proceedings of Transactions on Image Processing (TIP)}, 31:\penalty0 1072--1083, 2022{\natexlab{b}}.

\bibitem[Liu et~al.(2023{\natexlab{b}})Liu, Gupta, and Wang]{liu2023building}
Shaowei Liu, Saurabh Gupta, and Shenlong Wang.
\newblock Building rearticulable models for arbitrary 3d objects from 4d point clouds.
\newblock In \emph{Proceedings of Conference on Computer Vision and Pattern Recognition (CVPR)}, 2023{\natexlab{b}}.

\bibitem[Liu et~al.(2023{\natexlab{c}})Liu, Zhang, Hu, Huang, Wang, and Yi]{liu2023self}
Xueyi Liu, Ji~Zhang, Ruizhen Hu, Haibin Huang, He~Wang, and Li~Yi.
\newblock Self-supervised category-level articulated object pose estimation with part-level se (3) equivariance.
\newblock In \emph{Proceedings of International Conference on Learning Representations (ICLR)}, 2023{\natexlab{c}}.

\bibitem[Liu et~al.(2025)Liu, Jia, Chen, and Huang]{liu2025slotlifter}
Yu~Liu, Baoxiong Jia, Yixin Chen, and Siyuan Huang.
\newblock Slotlifter: Slot-guided feature lifting for learning object-centric radiance fields.
\newblock In \emph{Proceedings of European Conference on Computer Vision (ECCV)}, 2025.

\bibitem[Locatello et~al.(2020)Locatello, Weissenborn, Unterthiner, Mahendran, Heigold, Uszkoreit, Dosovitskiy, and Kipf]{locatello2020sa}
Francesco Locatello, Dirk Weissenborn, Thomas Unterthiner, Aravindh Mahendran, Georg Heigold, Jakob Uszkoreit, Alexey Dosovitskiy, and Thomas Kipf.
\newblock Object-centric learning with slot attention.
\newblock In \emph{Proceedings of Advances in Neural Information Processing Systems (NeurIPS)}, 2020.

\bibitem[Loper et~al.(2023)Loper, Mahmood, Romero, Pons-Moll, and Black]{loper2023smpl}
Matthew Loper, Naureen Mahmood, Javier Romero, Gerard Pons-Moll, and Michael~J Black.
\newblock Smpl: A skinned multi-person linear model.
\newblock In \emph{Seminal Graphics Papers: Pushing the Boundaries, Volume 2}, pp.\  851--866. 2023.

\bibitem[Lu et~al.(2024{\natexlab{a}})Lu, Zhang, Wang, Liu, Lu, and Tang]{lu2024manigaussian}
Guanxing Lu, Shiyi Zhang, Ziwei Wang, Changliu Liu, Jiwen Lu, and Yansong Tang.
\newblock Manigaussian: Dynamic gaussian splatting for multi-task robotic manipulation.
\newblock In \emph{Proceedings of European Conference on Computer Vision (ECCV)}, 2024{\natexlab{a}}.

\bibitem[Lu et~al.(2024{\natexlab{b}})Lu, Chen, Ni, Jia, Liu, Wan, Zeng, and Huang]{lu2024movis}
Ruijie Lu, Yixin Chen, Junfeng Ni, Baoxiong Jia, Yu~Liu, Diwen Wan, Gang Zeng, and Siyuan Huang.
\newblock Movis: Enhancing multi-object novel view synthesis for indoor scenes.
\newblock \emph{arXiv preprint arXiv:2412.11457}, 2024{\natexlab{b}}.

\bibitem[Lu et~al.(2024{\natexlab{c}})Lu, Guo, Hui, Chen, Yang, Tang, Zhu, and Dai]{lu20243d}
Zhicheng Lu, Xiang Guo, Le~Hui, Tianrui Chen, Min Yang, Xiao Tang, Feng Zhu, and Yuchao Dai.
\newblock 3d geometry-aware deformable gaussian splatting for dynamic view synthesis.
\newblock In \emph{Proceedings of Conference on Computer Vision and Pattern Recognition (CVPR)}, 2024{\natexlab{c}}.

\bibitem[Luiten et~al.(2024)Luiten, Kopanas, Leibe, and Ramanan]{luiten2024dynamic}
Jonathon Luiten, Georgios Kopanas, Bastian Leibe, and Deva Ramanan.
\newblock Dynamic 3d gaussians: Tracking by persistent dynamic view synthesis.
\newblock In \emph{Proceedings of International Conference on 3D Vision (3DV)}, 2024.

\bibitem[Luo et~al.(2025)Luo, Geng, Deng, Li, Wang, Jia, Guibas, and Huang]{luo2024physpart}
Rundong Luo, Haoran Geng, Congyue Deng, Puhao Li, Zan Wang, Baoxiong Jia, Leonidas Guibas, and Siyuang Huang.
\newblock Physpart: Physically plausible part completion for interactable objects.
\newblock In \emph{Proceedings of International Conference on Robotics and Automation (ICRA)}, 2025.

\bibitem[Ma et~al.(2023)Ma, Meng, Liu, Chen, Xu, and Chen]{ma2023sim2real}
Liqian Ma, Jiaojiao Meng, Shuntao Liu, Weihang Chen, Jing Xu, and Rui Chen.
\newblock Sim2real 2: Actively building explicit physics model for precise articulated object manipulation.
\newblock In \emph{Proceedings of International Conference on Robotics and Automation (ICRA)}, 2023.

\bibitem[Mandi et~al.(2024)Mandi, Weng, Bauer, and Song]{mandi2024real2code}
Zhao Mandi, Yijia Weng, Dominik Bauer, and Shuran Song.
\newblock Real2code: Reconstruct articulated objects via code generation.
\newblock \emph{arXiv preprint arXiv:2406.08474}, 2024.

\bibitem[Mao et~al.(2022)Mao, Zhang, Jiang, Chang, and Savva]{mao2022multiscan}
Yongsen Mao, Yiming Zhang, Hanxiao Jiang, Angel~X Chang, and Manolis Savva.
\newblock Multiscan: Scalable rgbd scanning for 3d environments with articulated objects.
\newblock In \emph{Proceedings of Conference on Computer Vision and Pattern Recognition (CVPR)}, 2022.

\bibitem[Mart{\'\i}n-Mart{\'\i}n et~al.(2016)Mart{\'\i}n-Mart{\'\i}n, H{\"o}fer, and Brock]{martin2016integrated}
Roberto Mart{\'\i}n-Mart{\'\i}n, Sebastian H{\"o}fer, and Oliver Brock.
\newblock An integrated approach to visual perception of articulated objects.
\newblock In \emph{Proceedings of International Conference on Robotics and Automation (ICRA)}, 2016.

\bibitem[Mihajlovic et~al.(2021)Mihajlovic, Zhang, Black, and Tang]{mihajlovic2021leap}
Marko Mihajlovic, Yan Zhang, Michael~J Black, and Siyu Tang.
\newblock Leap: Learning articulated occupancy of people.
\newblock In \emph{Proceedings of Conference on Computer Vision and Pattern Recognition (CVPR)}, 2021.

\bibitem[Mo et~al.(2021)Mo, Guibas, Mukadam, Gupta, and Tulsiani]{mo2021where2act}
Kaichun Mo, Leonidas~J Guibas, Mustafa Mukadam, Abhinav Gupta, and Shubham Tulsiani.
\newblock Where2act: From pixels to actions for articulated 3d objects.
\newblock In \emph{Proceedings of International Conference on Computer Vision (ICCV)}, 2021.

\bibitem[Mu et~al.(2021)Mu, Qiu, Kortylewski, Yuille, Vasconcelos, and Wang]{mu2021sdf}
Jiteng Mu, Weichao Qiu, Adam Kortylewski, Alan Yuille, Nuno Vasconcelos, and Xiaolong Wang.
\newblock A-sdf: Learning disentangled signed distance functions for articulated shape representation.
\newblock In \emph{Proceedings of International Conference on Computer Vision (ICCV)}, 2021.

\bibitem[Ni et~al.(2024)Ni, Chen, Jing, Jiang, Wang, Dai, Li, Zhu, Zhu, and Huang]{ni2024phyrecon}
Junfeng Ni, Yixin Chen, Bohan Jing, Nan Jiang, Bin Wang, Bo~Dai, Puhao Li, Yixin Zhu, Song-Chun Zhu, and Siyuan Huang.
\newblock Phyrecon: Physically plausible neural scene reconstruction.
\newblock 2024.

\bibitem[Nie et~al.(2022)Nie, Gadre, Ehsani, and Song]{nie2022structure}
Neil Nie, Samir~Yitzhak Gadre, Kiana Ehsani, and Shuran Song.
\newblock Structure from action: Learning interactions for articulated object 3d structure discovery.
\newblock \emph{arXiv preprint arXiv:2207.08997}, 2022.

\bibitem[Noguchi et~al.(2021)Noguchi, Sun, Lin, and Harada]{noguchi2021neural}
Atsuhiro Noguchi, Xiao Sun, Stephen Lin, and Tatsuya Harada.
\newblock Neural articulated radiance field.
\newblock In \emph{Proceedings of International Conference on Computer Vision (ICCV)}, 2021.

\bibitem[Noguchi et~al.(2022)Noguchi, Iqbal, Tremblay, Harada, and Gallo]{noguchi2022watch}
Atsuhiro Noguchi, Umar Iqbal, Jonathan Tremblay, Tatsuya Harada, and Orazio Gallo.
\newblock Watch it move: Unsupervised discovery of 3d joints for re-posing of articulated objects.
\newblock In \emph{Proceedings of Conference on Computer Vision and Pattern Recognition (CVPR)}, 2022.

\bibitem[Pillai et~al.(2015)Pillai, Walter, and Teller]{pillai2015learning}
Sudeep Pillai, Matthew~R Walter, and Seth Teller.
\newblock Learning articulated motions from visual demonstration.
\newblock \emph{arXiv preprint arXiv:1502.01659}, 2015.

\bibitem[Qian et~al.(2024)Qian, Wang, Mihajlovic, Geiger, and Tang]{qian20243dgs}
Zhiyin Qian, Shaofei Wang, Marko Mihajlovic, Andreas Geiger, and Siyu Tang.
\newblock 3dgs-avatar: Animatable avatars via deformable 3d gaussian splatting.
\newblock In \emph{Proceedings of Conference on Computer Vision and Pattern Recognition (CVPR)}, 2024.

\bibitem[Romero et~al.(2022)Romero, Tzionas, and Black]{romero2022embodied}
Javier Romero, Dimitrios Tzionas, and Michael~J Black.
\newblock Embodied hands: Modeling and capturing hands and bodies together.
\newblock \emph{arXiv preprint arXiv:2201.02610}, 2022.

\bibitem[Song et~al.(2023{\natexlab{a}})Song, Chen, Chen, Wei, Foo, Liu, and Lin]{song2023moda}
Chaoyue Song, Tianyi Chen, Yiwen Chen, Jiacheng Wei, Chuan~Sheng Foo, Fayao Liu, and Guosheng Lin.
\newblock Moda: Modeling deformable 3d objects from casual videos.
\newblock \emph{arXiv preprint arXiv:2304.08279}, 2023{\natexlab{a}}.

\bibitem[Song et~al.(2024)Song, Wei, Foo, Lin, and Liu]{song2024reacto}
Chaoyue Song, Jiacheng Wei, Chuan~Sheng Foo, Guosheng Lin, and Fayao Liu.
\newblock Reacto: Reconstructing articulated objects from a single video.
\newblock In \emph{Proceedings of Conference on Computer Vision and Pattern Recognition (CVPR)}, 2024.

\bibitem[Song et~al.(2023{\natexlab{b}})Song, Yang, Deng, Zhu, and Ramanan]{song2023total}
Chonghyuk Song, Gengshan Yang, Kangle Deng, Jun-Yan Zhu, and Deva Ramanan.
\newblock Total-recon: Deformable scene reconstruction for embodied view synthesis.
\newblock In \emph{Proceedings of International Conference on Computer Vision (ICCV)}, 2023{\natexlab{b}}.

\bibitem[Sturm et~al.(2011)Sturm, Stachniss, and Burgard]{sturm2011probabilistic}
J{\"u}rgen Sturm, Cyrill Stachniss, and Wolfram Burgard.
\newblock A probabilistic framework for learning kinematic models of articulated objects.
\newblock \emph{Journal of Artificial Intelligence Research}, 41, 2011.

\bibitem[Sun et~al.(2021)Sun, Shen, Wang, Bao, and Zhou]{sun2021loftr}
Jiaming Sun, Zehong Shen, Yuang Wang, Hujun Bao, and Xiaowei Zhou.
\newblock Loftr: Detector-free local feature matching with transformers.
\newblock In \emph{Proceedings of Conference on Computer Vision and Pattern Recognition (CVPR)}, 2021.

\bibitem[Sun et~al.(2023)Sun, Jiang, Savva, and Chang]{sun2023opdmulti}
Xiaohao Sun, Hanxiao Jiang, Manolis Savva, and Angel~Xuan Chang.
\newblock Opdmulti: Openable part detection for multiple objects.
\newblock \emph{arXiv preprint arXiv:2303.14087}, 2023.

\bibitem[Swaminathan et~al.(2024)Swaminathan, Gupta, Gupta, Maiya, Agarwal, and Shrivastava]{swaminathan2024leia}
Archana Swaminathan, Anubhav Gupta, Kamal Gupta, Shishira~R Maiya, Vatsal Agarwal, and Abhinav Shrivastava.
\newblock Leia: Latent view-invariant embeddings for implicit 3d articulation.
\newblock \emph{arXiv preprint arXiv:2409.06703}, 2024.

\bibitem[Tan et~al.(2023)Tan, Yang, and Ramanan]{tan2023distilling}
Jeff Tan, Gengshan Yang, and Deva Ramanan.
\newblock Distilling neural fields for real-time articulated shape reconstruction.
\newblock In \emph{Proceedings of Conference on Computer Vision and Pattern Recognition (CVPR)}, 2023.

\bibitem[Torne et~al.(2024)Torne, Simeonov, Li, Chan, Chen, Gupta, and Agrawal]{torne2024reconciling}
Marcel Torne, Anthony Simeonov, Zechu Li, April Chan, Tao Chen, Abhishek Gupta, and Pulkit Agrawal.
\newblock Reconciling reality through simulation: A real-to-sim-to-real approach for robust manipulation.
\newblock \emph{arXiv preprint arXiv:2403.03949}, 2024.

\bibitem[Tseng et~al.(2022)Tseng, Liao, Yen-Chen, and Sun]{tseng2022cla}
Wei-Cheng Tseng, Hung-Ju Liao, Lin Yen-Chen, and Min Sun.
\newblock Cla-nerf: Category-level articulated neural radiance field.
\newblock In \emph{Proceedings of International Conference on Robotics and Automation (ICRA)}, 2022.

\bibitem[Wan et~al.(2024{\natexlab{a}})Wan, Lu, and Zeng]{wan2024superpoint}
Diwen Wan, Ruijie Lu, and Gang Zeng.
\newblock Superpoint gaussian splatting for real-time high-fidelity dynamic scene reconstruction.
\newblock \emph{arXiv preprint arXiv:2406.03697}, 2024{\natexlab{a}}.

\bibitem[Wan et~al.(2024{\natexlab{b}})Wan, Wang, Lu, and Zeng]{wan2024template}
Diwen Wan, Yuxiang Wang, Ruijie Lu, and Gang Zeng.
\newblock Template-free articulated gaussian splatting for real-time reposable dynamic view synthesis.
\newblock \emph{arXiv preprint arXiv:2412.05570}, 2024{\natexlab{b}}.

\bibitem[Wang et~al.(2021)Wang, Liu, Liu, Theobalt, Komura, and Wang]{wang2021neus}
Peng Wang, Lingjie Liu, Yuan Liu, Christian Theobalt, Taku Komura, and Wenping Wang.
\newblock Neus: Learning neural implicit surfaces by volume rendering for multi-view reconstruction.
\newblock \emph{arXiv preprint arXiv:2106.10689}, 2021.

\bibitem[Wang et~al.(2019)Wang, Zhou, Shi, Chen, Zhao, and Xu]{wang2019shape2motion}
Xiaogang Wang, Bin Zhou, Yahao Shi, Xiaowu Chen, Qinping Zhao, and Kai Xu.
\newblock Shape2motion: Joint analysis of motion parts and attributes from 3d shapes.
\newblock In \emph{Proceedings of Conference on Computer Vision and Pattern Recognition (CVPR)}, 2019.

\bibitem[Wang et~al.(2004)Wang, Bovik, Sheikh, and Simoncelli]{wang2004image}
Zhou Wang, Alan~C Bovik, Hamid~R Sheikh, and Eero~P Simoncelli.
\newblock Image quality assessment: from error visibility to structural similarity.
\newblock \emph{Proceedings of Transactions on Image Processing (TIP)}, 13\penalty0 (4):\penalty0 600--612, 2004.

\bibitem[Wei et~al.(2022)Wei, Chabra, Ma, Lassner, Zollh{\"o}fer, Rusinkiewicz, Sweeney, Newcombe, and Slavcheva]{wei2022self}
Fangyin Wei, Rohan Chabra, Lingni Ma, Christoph Lassner, Michael Zollh{\"o}fer, Szymon Rusinkiewicz, Chris Sweeney, Richard Newcombe, and Mira Slavcheva.
\newblock Self-supervised neural articulated shape and appearance models.
\newblock In \emph{Proceedings of Conference on Computer Vision and Pattern Recognition (CVPR)}, 2022.

\bibitem[Wen et~al.(2023)Wen, Tremblay, Blukis, Tyree, M{\"u}ller, Evans, Fox, Kautz, and Birchfield]{wen2023bundlesdf}
Bowen Wen, Jonathan Tremblay, Valts Blukis, Stephen Tyree, Thomas M{\"u}ller, Alex Evans, Dieter Fox, Jan Kautz, and Stan Birchfield.
\newblock Bundlesdf: Neural 6-dof tracking and 3d reconstruction of unknown objects.
\newblock In \emph{Proceedings of Conference on Computer Vision and Pattern Recognition (CVPR)}, 2023.

\bibitem[Weng et~al.(2021)Weng, Wang, Zhou, Qin, Duan, Fan, Chen, Su, and Guibas]{weng2021captra}
Yijia Weng, He~Wang, Qiang Zhou, Yuzhe Qin, Yueqi Duan, Qingnan Fan, Baoquan Chen, Hao Su, and Leonidas~J Guibas.
\newblock Captra: Category-level pose tracking for rigid and articulated objects from point clouds.
\newblock In \emph{Proceedings of International Conference on Computer Vision (ICCV)}, 2021.

\bibitem[Weng et~al.(2024)Weng, Wen, Tremblay, Blukis, Fox, Guibas, and Birchfield]{weng2024neural}
Yijia Weng, Bowen Wen, Jonathan Tremblay, Valts Blukis, Dieter Fox, Leonidas Guibas, and Stan Birchfield.
\newblock Neural implicit representation for building digital twins of unknown articulated objects.
\newblock In \emph{Proceedings of Conference on Computer Vision and Pattern Recognition (CVPR)}, 2024.

\bibitem[Wu et~al.(2024)Wu, Yi, Fang, Xie, Zhang, Wei, Liu, Tian, and Wang]{wu20244d}
Guanjun Wu, Taoran Yi, Jiemin Fang, Lingxi Xie, Xiaopeng Zhang, Wei Wei, Wenyu Liu, Qi~Tian, and Xinggang Wang.
\newblock 4d gaussian splatting for real-time dynamic scene rendering.
\newblock In \emph{Proceedings of Conference on Computer Vision and Pattern Recognition (CVPR)}, 2024.

\bibitem[Xiang et~al.(2020)Xiang, Qin, Mo, Xia, Zhu, Liu, Liu, Jiang, Yuan, Wang, et~al.]{xiang2020sapien}
Fanbo Xiang, Yuzhe Qin, Kaichun Mo, Yikuan Xia, Hao Zhu, Fangchen Liu, Minghua Liu, Hanxiao Jiang, Yifu Yuan, He~Wang, et~al.
\newblock Sapien: A simulated part-based interactive environment.
\newblock In \emph{Proceedings of Conference on Computer Vision and Pattern Recognition (CVPR)}, 2020.

\bibitem[Xie et~al.(2024)Xie, Zong, Qiu, Li, Feng, Yang, and Jiang]{xie2024physgaussian}
Tianyi Xie, Zeshun Zong, Yuxing Qiu, Xuan Li, Yutao Feng, Yin Yang, and Chenfanfu Jiang.
\newblock Physgaussian: Physics-integrated 3d gaussians for generative dynamics.
\newblock In \emph{Proceedings of Conference on Computer Vision and Pattern Recognition (CVPR)}, 2024.

\bibitem[Xu et~al.(2020)Xu, Bazavan, Zanfir, Freeman, Sukthankar, and Sminchisescu]{xu2020ghum}
Hongyi Xu, Eduard~Gabriel Bazavan, Andrei Zanfir, William~T Freeman, Rahul Sukthankar, and Cristian Sminchisescu.
\newblock Ghum \& ghuml: Generative 3d human shape and articulated pose models.
\newblock In \emph{Proceedings of Conference on Computer Vision and Pattern Recognition (CVPR)}, 2020.

\bibitem[Yan et~al.(2020)Yan, Hu, Yan, Chen, Van~Kaick, Zhang, and Huang]{yan2020rpm}
Zihao Yan, Ruizhen Hu, Xingguang Yan, Luanmin Chen, Oliver Van~Kaick, Hao Zhang, and Hui Huang.
\newblock Rpm-net: recurrent prediction of motion and parts from point cloud.
\newblock \emph{arXiv preprint arXiv:2006.14865}, 2020.

\bibitem[Yang et~al.(2024{\natexlab{a}})Yang, Chen, He, Cai, Yang, Wu, and Lin]{yang2024attrihuman}
Fan Yang, Tianyi Chen, Xiaosheng He, Zhongang Cai, Lei Yang, Si~Wu, and Guosheng Lin.
\newblock Attrihuman-3d: Editable 3d human avatar generation with attribute decomposition and indexing.
\newblock In \emph{Proceedings of Conference on Computer Vision and Pattern Recognition (CVPR)}, 2024{\natexlab{a}}.

\bibitem[Yang et~al.(2021{\natexlab{a}})Yang, Sun, Jampani, Vlasic, Cole, Chang, Ramanan, Freeman, and Liu]{yang2021lasr}
Gengshan Yang, Deqing Sun, Varun Jampani, Daniel Vlasic, Forrester Cole, Huiwen Chang, Deva Ramanan, William~T Freeman, and Ce~Liu.
\newblock Lasr: Learning articulated shape reconstruction from a monocular video.
\newblock In \emph{Proceedings of Conference on Computer Vision and Pattern Recognition (CVPR)}, 2021{\natexlab{a}}.

\bibitem[Yang et~al.(2021{\natexlab{b}})Yang, Sun, Jampani, Vlasic, Cole, Liu, and Ramanan]{yang2021viser}
Gengshan Yang, Deqing Sun, Varun Jampani, Daniel Vlasic, Forrester Cole, Ce~Liu, and Deva Ramanan.
\newblock Viser: Video-specific surface embeddings for articulated 3d shape reconstruction.
\newblock \emph{Proceedings of Advances in Neural Information Processing Systems (NeurIPS)}, 2021{\natexlab{b}}.

\bibitem[Yang et~al.(2022)Yang, Vo, Neverova, Ramanan, Vedaldi, and Joo]{yang2022banmo}
Gengshan Yang, Minh Vo, Natalia Neverova, Deva Ramanan, Andrea Vedaldi, and Hanbyul Joo.
\newblock Banmo: Building animatable 3d neural models from many casual videos.
\newblock In \emph{Proceedings of Conference on Computer Vision and Pattern Recognition (CVPR)}, 2022.

\bibitem[Yang et~al.(2023{\natexlab{a}})Yang, Wang, Reddy, and Ramanan]{yang2023reconstructing}
Gengshan Yang, Chaoyang Wang, N~Dinesh Reddy, and Deva Ramanan.
\newblock Reconstructing animatable categories from videos.
\newblock In \emph{Proceedings of Conference on Computer Vision and Pattern Recognition (CVPR)}, 2023{\natexlab{a}}.

\bibitem[Yang et~al.(2023{\natexlab{b}})Yang, Yang, Zhang, Manchester, and Ramanan]{yang2023ppr}
Gengshan Yang, Shuo Yang, John~Z Zhang, Zachary Manchester, and Deva Ramanan.
\newblock Ppr: Physically plausible reconstruction from monocular videos.
\newblock In \emph{Proceedings of International Conference on Computer Vision (ICCV)}, 2023{\natexlab{b}}.

\bibitem[Yang et~al.(2024{\natexlab{b}})Yang, Jia, Zhi, and Huang]{yang2024physcene}
Yandan Yang, Baoxiong Jia, Peiyuan Zhi, and Siyuan Huang.
\newblock Physcene: Physically interactable 3d scene synthesis for embodied ai.
\newblock In \emph{Proceedings of Conference on Computer Vision and Pattern Recognition (CVPR)}, 2024{\natexlab{b}}.

\bibitem[Yariv et~al.(2021)Yariv, Gu, Kasten, and Lipman]{yariv2021volume}
Lior Yariv, Jiatao Gu, Yoni Kasten, and Yaron Lipman.
\newblock Volume rendering of neural implicit surfaces.
\newblock \emph{Proceedings of Advances in Neural Information Processing Systems (NeurIPS)}, 2021.

\bibitem[Yi et~al.(2018)Yi, Huang, Liu, Kalogerakis, Su, and Guibas]{yi2018deep}
Li~Yi, Haibin Huang, Difan Liu, Evangelos Kalogerakis, Hao Su, and Leonidas Guibas.
\newblock Deep part induction from articulated object pairs.
\newblock \emph{arXiv preprint arXiv:1809.07417}, 2018.

\bibitem[Zhang et~al.(2021)Zhang, Litany, Sridhar, and Guibas]{zhang2021strobenet}
Ge~Zhang, Or~Litany, Srinath Sridhar, and Leonidas Guibas.
\newblock Strobenet: Category-level multiview reconstruction of articulated objects.
\newblock \emph{arXiv preprint arXiv:2105.08016}, 2021.

\bibitem[Zhang et~al.(2018)Zhang, Isola, Efros, Shechtman, and Wang]{zhang2018unreasonable}
Richard Zhang, Phillip Isola, Alexei~A Efros, Eli Shechtman, and Oliver Wang.
\newblock The unreasonable effectiveness of deep features as a perceptual metric.
\newblock In \emph{Proceedings of Conference on Computer Vision and Pattern Recognition (CVPR)}, 2018.

\bibitem[Zhao et~al.(2024)Zhao, Li, Li, Qi, Ruan, Zhu, and Althoefer]{zhao2024tac}
Zihang Zhao, Yuyang Li, Wanlin Li, Zhenghao Qi, Lecheng Ruan, Yixin Zhu, and Kaspar Althoefer.
\newblock Tac-man: Tactile-informed prior-free manipulation of articulated objects.
\newblock \emph{Transactions on Robotics (T-RO)}, 2024.

\bibitem[Zhou et~al.(2018)Zhou, Park, and Koltun]{zhou2018open3d}
Qian-Yi Zhou, Jaesik Park, and Vladlen Koltun.
\newblock Open3d: A modern library for 3d data processing.
\newblock \emph{arXiv preprint arXiv:1801.09847}, 2018.

\bibitem[Zhu et~al.(2024)Zhu, Zhang, Ma, Niu, Chen, Jia, Deng, Huang, and Li]{zhu2024unifying}
Ziyu Zhu, Zhuofan Zhang, Xiaojian Ma, Xuesong Niu, Yixin Chen, Baoxiong Jia, Zhidong Deng, Siyuan Huang, and Qing Li.
\newblock Unifying 3d vision-language understanding via promptable queries.
\newblock In \emph{Proceedings of European Conference on Computer Vision (ECCV)}, 2024.

\bibitem[Zuffi et~al.(2017)Zuffi, Kanazawa, Jacobs, and Black]{zuffi20173d}
Silvia Zuffi, Angjoo Kanazawa, David~W Jacobs, and Michael~J Black.
\newblock 3d menagerie: Modeling the 3d shape and pose of animals.
\newblock In \emph{Proceedings of Conference on Computer Vision and Pattern Recognition (CVPR)}, 2017.

\end{thebibliography}
\bibliographystyle{iclr2025_conference}

\clearpage

\appendix

\renewcommand{\thefigure}{A.\arabic{figure}}
\renewcommand{\thetable}{A.\arabic{table}}
\renewcommand{\theequation}{A.\arabic{equation}}
\setcounter{section}{0}
\setcounter{figure}{0}
\setcounter{table}{0}
\setcounter{equation}{0}

\section{Implementation and Training Details}
\label{app:imp}
\paragraph{Canonical Gaussian Initialzation}
We train single-state Gaussians $\mathcal{G}^0$ and $\mathcal{G}^1$ for 10K steps with loss $\mathcal{L}=(1-\lambda_\text{{SSIM}})\mathcal{L}_I+\lambda_\text{{SSIM}}\mathcal{L}_{\text{D-SSIM}}+\lambda_o\mathcal{L}_o$, where $\lambda_\text{{SSIM}}=0.2,\lambda_o=0.01$ is used in experiments and $\mathcal{L}_o$ is an opacity entropy loss calculated as:
$$
\hat{\sigma}_i=\mathbb{1}\{\sigma_i>0.5\},\quad
\mathcal{L}_o=-\frac{1}{N}\sum_{i=1}^{N}[\hat{\sigma}_i\sigma_i+(1-\hat{\sigma}_i)\log(1-\sigma_i)],
$$ 
which encourages Gaussian opacities $\sigma_i$ to approach either 0 or 1, controlling Gaussian count and accelerating training. We then obtain coarse canonical Gaussians by matching $\mathcal{G}^0$ and $\mathcal{G}^1$ as described in \cref{sec:method:canonical}. This stage takes about 2 minutes per object.
\paragraph{Part Discovery for Articulation Modeling} 
\label{app:imp:assignment}
As described in \cref{sec:method:skinning}, given canonical Gaussians $\mathcal{G}^c = \{G_i\}_{i=1}^{N}$ and $K$ learnable part centers $C_k = (\vp_k, \mR_k, {\bm \lambda}_k)$, we calculate part-level masks $\mM$ using \cref{eq:part_assignment}. We use a learnable hash grid $H$ to encode Gaussian positions and predict the residual term in \cref{eq:part_assignment} as:
\begin{equation*}
\begin{aligned}
\rmX^k_i = \frac{[\mR_k(\vmu^c_i-\vp_k)]}{{\bm\lambda}_k}, &\quad
 \rmD_{ik} = (\rmX^k_i)^T\cdot\rmX^k_i \\
{\mW_\Delta}_{ik}=\mathrm{MLP}(\vmu^c_i,H(\vmu^c_i),\{X^k_i\}_{k=1}^K,\{D_{ik}\}_{k=1}^K), &\quad\bm{M} = \mathrm{GumbelSoftmax}\left(\frac{-\mD + \mW_{\Delta}}{\tau}\right) \\
\end{aligned}
\end{equation*}
Since the part assignment and articulation parameters are far from optimal at the beginning of training, using hard assignment for Gumbel-Softmax hinders the joint optimization of the part assignment and articulation parameters. To address this problem, we anneal the temperature $\tau$ from 1 to 0.1 over 10K steps, using soft assignment that is similar to Softmax when $\tau > 0.1$ and hard assignment otherwise for training stability. This approach allows for more flexible assignments during the early stages of training, facilitating better joint optimization, and gradually transitioning to decisive part assignment as the model converges.

\paragraph{Optimization}
To enhance the learning of articulation parameters, we adopt a warm-up strategy to predict the joint type of each part. This process requires 3K-5K steps that take 30 to 50 seconds. Then we train \model with joint type constraint for 20K steps, taking 5-7 minutes per object. 
For hyper-parameters, we set the threshold $\epsilon_{\text{static}}$ to identify static/movable Gaussians as $\epsilon_{\text{static}}=0.02 \cdot \max_{i} \text{CD}_{i}^{t\rightarrow \bar{t}}$ for two-part objects and $\epsilon_{\text{static}}=0.05 \cdot \max_{i} \text{CD}_{i}^{t\rightarrow \bar{t}}$ for multi-part objects. We use $\epsilon_{\text{revol}}=10^\degree$ for predicting joint types following PARIS~\citep{jiayi2023paris}.
$\lambda_{cd}$ and $\lambda_{reg}$ are set as 100 and 0.1 separately.
In addition, the CD loss in \cref{eq:cd_loss} aims to decrease the distance between a deformed Gaussian $G_i^t$ and its nearest Gaussians $\hat{G}_i^t$ in $\mathcal{G}^t_{\text{single}}$. Since the deformed Gaussians and canonical Gaussians for a prismatic joint have a large overlap, the nearest Gaussian may be in the opposite direction of the ideal one, making it ineffective for prismatic joints. Thus the CD loss is only used for regularizing the objects that only have revolute joints.
Moreover, the densification strategy of Gaussians is cloning or splitting one Gaussian when the gradient of its center $\vmu$ is greater than a threshold $\epsilon_{\text{densify}}$. This is effective for static scenes but meets challenges for dynamic scenes. In the early stage of training, the large gradient is often due to deformation error. To prevent excessive increase of Gaussian quantity due to deformation error, we raised this threshold $\epsilon_{\text{densify}}$ from 0.0002 used in previous works \citep{kerbl20233d,huang2024sc} to 0.001.

\section{Additional Discussions}
\label{app:discussion}
\begin{table*}[t]
\caption{\textbf{Quantitative evaluation of each state on PARIS data.} We report the average of metrics over 10 trials of each state. "metric-0/1" represents the metric evaluated at state 0/1 and "metric-m" is the average of two states. We highlight \colorbox[HTML]{ffc5c5}{best} results on average of two states. Axis Pos. is omitted for prismatic joints (Blade, Storage, and Real Storage).}
\label{tab:app:exp_2part}
\renewcommand{\arraystretch}{1.2}
\resizebox{\linewidth}{!}{
\begin{tabular}{cc|ccccccccccc|ccc}
\hline
\multirow{2}{*}{Metric} &\multirow{2}{*}{Method} &\multicolumn{11}{c}{Synthetic Objects} &\multicolumn{3}{|c}{Real Objects} \\
& &FoldChair &Fridge &Laptop &Oven &Scissor &Stapler &USB 
&Washer&Blade &Storage &All & Fridge &Storage &All \\
\hline
\multirow{6}{*}{\shortstack{Axis\\Ang}} 
&DTA-0
&0.03 &0.09 &0.07 &0.22 &0.10 &0.06 &0.11 &0.36 &0.20 &0.07 &0.13 &2.08 &13.64 &7.86 \\
&Ours-0
&0.01 &0.03 &0.01 &0.01 &0.05 &0.01 &0.04 &0.02 &0.03 &0.01 &0.02 & 2.09 &3.47 &2.78 \\
&DTA-1
&0.04 &0.10 &0.07 &0.23 &0.10 &0.07 &0.11 &0.36 &0.26 &0.09 &0.14 &2.07 &8.08 &5.08 \\
&Ours-1
&0.01 &0.03 &0.01 &0.01 &0.05 &0.01 &0.04 &0.02 &0.03 &0.01 &0.02 & 2.09 &3.47 &2.78 \\
&DTA-m
&0.04 &0.10 &0.07 &0.22 &0.10 &0.06 &0.11 &0.36 &0.23 &0.08 &0.14 &\best{2.08} &10.86 &6.47 \\
&Ours-m
&\best{0.01} &\best{0.03} &\best{0.01} &\best{0.01} &\best{0.05} &\best{0.01} &\best{0.04} &\best{0.02} &\best{0.03} &\best{0.01} &\best{0.02} & 2.09 &\best{3.47} &\best{2.78} \\
\hline
\multirow{6}{*}{\shortstack{Axis\\Pos}} 
&DTA-0
&0.01 &0.01 &0.01 &0.01 &0.03 &0.02 &0.00 &0.04 &- &- &0.02 &0.59 &- &0.59 \\
&Ours-0 
&0.00 &0.00 &0.01 &0.00 &0.00 &0.01 &0.00 &0.00 & - & - &0.00 &0.47 & - &0.47 \\
&DTA-1
&0.01 &0.01 &0.01 &0.01 &0.02 &0.02 &0.00 &0.05 &- &- &0.02 &0.59 &- &0.59 \\
&Ours-1
&0.00 &0.00 &0.01 &0.00 &0.00 &0.01 &0.00 &0.00 & - & - &0.00 &0.47 & - &0.47 \\
&DTA-m
&0.01 &0.01 &0.01 &0.01 &0.03 &0.02 &0.00 &0.04 &- &- &0.02 &0.59 &- &0.59 \\
&Ours-m
&\best{0.00} &\best{0.00} &\best{0.01} &\best{0.00} &\best{0.00} &\best{0.01} &\best{0.00} &\best{0.00} & \best{-} & \best{-} &\best{0.00} &\best{0.47} & \best{-} &\best{0.47} \\
\hline
\multirow{6}{*}{\shortstack{Part\\Motion}}
&DTA-0
&0.10 &0.12 &0.11 &0.12 &0.38 &0.08 &0.15 &0.28 &0.00 &0.00 &0.13 &1.85 &0.14 &1.00 \\
&Ours-0
&0.03 &0.04
&0.02 &0.02
&0.04 &0.01
&0.03 &0.03
&0.00 &0.00
&0.02 & 1.94
&0.04 &0.99 \\
&DTA-1
&0.09 &0.13 &0.11 &0.13 &0.37 &0.08 &0.14 &0.28 &0.00 &0.00 &0.13 &1.85 &0.09 &0.97 \\
&Ours-1
&0.03 &0.04
&0.02 &0.02
&0.04 &0.01
&0.03 &0.03
&0.00 &0.00
&0.02 & 1.94
&0.04 &0.99 \\
&DTA-m
&0.09 &0.12 &0.11 &0.12 &0.38 &0.08 &0.15 &0.28 &0.00 &0.00 &0.13 &\best{1.85} &0.12 &0.99 \\
&Ours-m
&\best{0.03} &\best{0.04}
&\best{0.02} &\best{0.02}
&\best{0.04} &\best{0.01}
&\best{0.03} &\best{0.03}
&\best{0.00} &\best{0.00}
&\best{0.02} & 1.94
&\best{0.04} &\best{0.99} \\
\hline
\multirow{6}{*}{\shortstack{CD-s}} 
&DTA-0
&0.18 &0.62 &0.32 &4.60 &3.30 &2.68 &2.32 &4.77 &0.55 &4.71 &2.41 &2.36 &10.98 &6.67 \\
&Ours-0
&0.26 & 0.52 & 0.59 & 3.88 & 0.62 & 3.85 & 2.25 & 6.41 & 0.54 & 7.47 & 2.64 & 1.64 & 2.93 & 2.29 \\
&DTA-1
&0.19 &0.63 &0.30 &4.58 &3.55 &2.91 &2.90 &4.56 &0.45 &4.90 &2.50 &2.59 &9.60 &6.10 \\
&Ours-1
&0.26 & 0.48 & 0.63 & 4.00 & 0.61 & 3.83 & 2.56 & 6.43 & 0.54 & 7.31 & 2.67 & 2.01 & 4.02 & 3.02 \\
&DTA-m
&\best{0.19} &0.62 &\best{0.31} &4.59 &3.43 &\best{2.79} &2.61 &\best{4.66} &\best{0.50} &\best{4.80} &\best{2.46} &2.48 &10.29 &6.39 \\
&Ours-m
&0.26 &\best{0.50} 
&0.61 &\best{3.94} 
&\best{0.61} &3.84 
&\best{2.41} &6.42 &0.54 &7.39
&2.65 &\best{1.82 } 
&\best{3.48} &\best{2.65} \\
\hline
\multirow{6}{*}{\shortstack{CD-m}}
&DTA-0
&0.15 &0.27 &0.16 &0.44 &17.38 &2.34 &1.47 &0.37 &2.05 &0.36 &2.50 &1.12 &30.78 &15.95 \\
&Ours-0
&0.54 & 0.21 & 0.14 & 0.89 & 0.65 & 0.88 & 1.22 & 1.54 & 1.12 & 1.03 & 0.82 & 0.66 & 6.28 & 3.47 \\
&DTA-1
&0.13 &0.30 &0.13 &0.45 &10.11 &1.13 &1.51 &0.45 &61.38 &0.36 &7.60 &1.85 &365.74 &183.80 \\
&Ours-1
&0.12 & 0.21 & 0.13 & 0.76 & 0.64 & 0.52 & 1.43 & 0.45 & 1.01 & 1.02 & 0.63 & 1.31 & 87.81 & 44.56 \\
&DTA-m
&\best{0.14} &0.28 &0.15 &\best{0.44} &13.75 &1.73 &1.49 &\best{0.41} &31.72 &\best{0.36} &5.05 &1.48 &198.26 &99.88 \\
&Ours-m
&0.33 &\best{0.21} &\best{0.14} &0.82 &\best{0.65} &\best{0.70} &\best{1.33} &1.00
&\best{1.06} &1.02 &\best{0.73} &\best{0.99} &\best{47.05} &\best{24.02} \\

\hline
\multirow{6}{*}{\shortstack{CD-w}}
&DTA-0
&0.27 &0.70 &0.35 &4.24 &0.42 &2.13 &1.17 &4.59 &0.36 &4.09 &1.83 &2.08 &8.98 &5.53 \\
&Ours-0
&0.43 & 0.58 & 0.47 & 3.58 & 0.69 & 3.13 & 1.28 & 6.12 & 0.61 & 5.13 & 2.20 & 1.29 & 3.23 & 2.26 \\
&DTA-1
&0.26 &0.70 &0.32 &4.27 &0.41 &1.92 &1.52 &4.48 &0.38 &3.99 &1.83 &2.19 &9.03 &5.61 \\
&Ours-1
&0.30 & 0.59 & 0.50 & 3.71 & 0.67 & 2.63 & 1.87 & 5.99 & 0.65 & 5.21 & 2.21 & 1.45 & 2.45 & 1.95 \\
&DTA-m
&\best{0.26} &0.70 &\best{0.34} &4.25 &\best{0.41} &\best{2.02} &\best{1.34} &\best{4.53} &\best{0.37} &\best{4.04} &\best{1.83} &2.13 &9.01 &5.57 \\
&Ours-m
&0.36 &\best{0.59} &0.48 &\best{3.64} &0.68 &2.88 & 1.58 &6.05 & 0.63 & 5.17 &{2.21} &\best{1.37} &\best{2.84} &\best{2.11} \\

\hline
\end{tabular}
}
\end{table*}

We present a comprehensive analysis of \model and DTA through additional quantitative and qualitative results. 

\paragraph{Visibility Problem} Our results uncover an intriguing inconsistency in DTA's performance across different states of the same object. As illustrated in \cref{tab:app:exp_2part}, DTA demonstrates good reconstruction quality in the high-visibility state but shows markedly poor performance in the low-visibility state. This limitation is particularly pronounced for objects with prismatic joints, such as real storage and blade. In these cases, DTA struggles to accurately capture the geometry and articulation of partially occluded parts. 
The observed inconsistency and state-dependent performance fluctuations underscore the necessity for a more robust approach that effectively connects and leverages information from multiple states. This is precisely where \model's strengths become evident. By establishing connections between different articulation states, \model achieves more consistent and high-quality reconstructions across varying object configurations.
Jointly optimizing over multiple states allows \model to:
1) Leverage complementary information from different articulation states,
2) Maintain consistency in part assignment and geometry across states,
3) Better handle occlusions and low-visibility scenarios by inferring occluded geometries from other states.
These capabilities enable \model to produce more accurate and reliable reconstructions, particularly in challenging scenarios. The superior performance of \model demonstrates its potential for robust articulated object reconstruction in real-world applications.

\begin{figure}[t!]
 \centering
 \resizebox{\linewidth}{!}{\includegraphics[width=\linewidth]{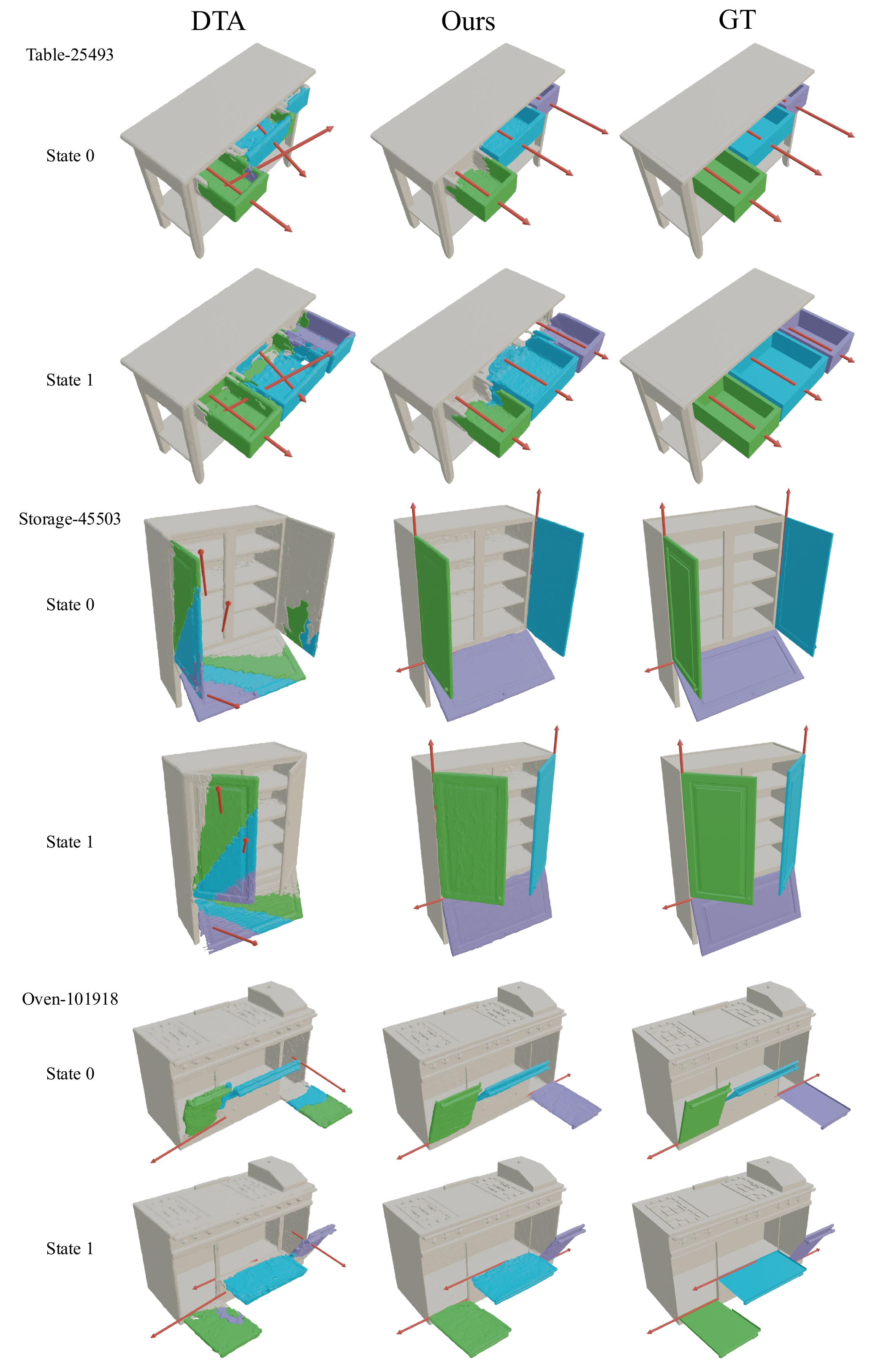}}
 \caption{\textbf{Additional qualitative results on \model-Multi.}}
 \label{fig:app:mpart}
 \vspace{-10pt}
\end{figure}

\paragraph{Significance of Part Assignment}
Through analysis of both qualitative (\cref{fig:app:mpart}) and quantitative (\cref{tab:exp_mpart_our}) results, we have identified that the model's ultimate performance is primarily determined by the accuracy of part assignment. When the model fails to correctly divide an object into parts, it becomes impossible to obtain reasonable joint parameter estimation. Conversely, even when joint parameter estimation is inaccurate, the model may still correctly separate the object's parts. This insight reveals that accurate part assignment is a crucial prerequisite for high-quality articulated object reconstruction.
Our findings emphasize that to enhance the reconstruction of articulated objects, the ability to reasonably separate parts is of paramount importance. \model addresses this challenge through the center-based segmentation and improved initialization by clustering. These techniques work in synergy to significantly improve the part segmentation capabilities of \model. By enhancing the model's ability to correctly identify and separate object parts, we lay a solid foundation for subsequent stages of the reconstruction process, including joint parameter estimation and final mesh reconstruction.

\section{Limitations}
\label{app:limitation}
\paragraph{Stability of Randomness.} 
\model exhibits enhanced robustness and stability across different random seeds, primarily due to our innovative initialization strategy for canonical Gaussians and our part assignment module. We observe that stability issues often stem from the initialization of three key components: canonical Gaussians $\mathcal{G}^c$, part centers $C$ in the part assignment module, and joint articulation parameters $\Psi$.
As demonstrated in \cref{sec:exp:ablation}, faulty initialization of $\mathcal{G}^c$ and $C$ can lead to significant performance degradation, particularly for complex objects with multiple movable parts. While our current initialization strategy has greatly improved stability, severe initialization errors in center $C$ may still result in part mis-segmentation. We can integrate prior models such as SAM~\citep{kirillov2023segment} to enhance the ability to correct center initialization errors.
Although \model works with randomly initialized $\Psi$, we have observed that improved initialization of $\Psi$ brings enhanced performance. Future work could explore the integration of heuristic algorithms or feed-forward articulation estimation models to provide better initial estimation for $\Psi$.

\paragraph{Limited States} 
Our current approach is limited to modeling articulated objects using only two states, which may not fully capture the complexity of real-world multi-part objects. 
Moreover, as the number of parts increases, distinguishing parts with similar joint axes and motion patterns (such as parallel drawers) becomes increasingly challenging, complicating the segmentation process.
To address this, two main avenues could be explored for future research:
1) Multi-state Extension: Develop a methodology to extend \model to handle multiple states that interact with different parts, potentially by identifying movable parts with a sequential state update mechanism. This would involve iteratively updating the model as new state information becomes available, allowing for a more comprehensive representation of the object's articulation space.
2) Continuous Temporal Reconstruction: Adapt \model to reconstruct articulated objects from monocular video sequences. This approach would leverage temporal information to infer a continuous range of articulation states, providing a more nuanced understanding of the object's movement capabilities.

\paragraph{Mesh Reconstruction Fidelity} 
Our current implementation utilizes the original Gaussian Splatting technique, which, while effective, has limitations in terms of mesh reconstruction quality compared with NeRF-based methods\citep{wang2021neus,yariv2021volume,wen2023bundlesdf}. Integrating recent advancements in reconstruction with Gaussian Splatting~\citep{huang20242d,chen2024pgsr} may help to improve the reconstruction fidelity of \model.

{\section{Additional Experiments}}
{\subsection{Additional Quantitative Comparisons}}
\label{app:addi_ac}
\begin{table*}[t]
\caption{{\textbf{Quantitative evaluation of Axis Pos metric on PARIS.} Metrics are reported as mean $\pm$ std over 10 trials on average of 2 states. We report the value timed by 1000 and highlight the \colorbox[HTML]{ffc5c5}{best} results.}}
\label{tab:app:exp_ap}
\renewcommand{\arraystretch}{1.2}
\resizebox{\linewidth}{!}{
{\begin{tabular}{cc|ccccccccc}
\hline
{Metric} &{Method} &FoldChair &Fridge &Laptop &Oven &Scissor &Stapler &USB 
&Washer &All \\
\hline
\multirow{2}{*}{\shortstack{Axis\\Pos}} 
&DTA
&{0.53\tiny{$\pm$0.3}} &{0.62\tiny{$\pm$0.3}} &{1.10\tiny{$\pm$0.7}}
&{1.49\tiny{$\pm$1.0}} &{2.48\tiny{$\pm$2.8}} &{2.21\tiny{$\pm$1.8}} &{0.35\tiny{$\pm$0.2}} &{4.53\tiny{$\pm$2.8}}  &{1.66\tiny{$\pm$}1.2}  \\
&Ours
&\best{0.48\tiny{$\pm$0.2}} &\best{0.44\tiny{$\pm$0.2}} &\best{0.39\tiny{$\pm$0.3}} &\best{0.55\tiny{$\pm$0.4}} &\best{0.16\tiny{$\pm$0.1}} &\best{0.93\tiny{$\pm$0.4}} &\best{0.08\tiny{$\pm$0.1}} &\best{0.33\tiny{$\pm$0.3}} &\best{0.42\tiny{$\pm$0.3}}  \\
\hline
\end{tabular}
}
}
\end{table*}

\begin{table*}[t]
\caption{{\textbf{Quantitative evaluation for perception-based metrics on PARIS data.} We report the results on average of two states. We highlight \colorbox[HTML]{ffc5c5}{best} results.}}
\label{tab:app:exp_psnr}
\renewcommand{\arraystretch}{1.2}
\resizebox{\linewidth}{!}{
{
\begin{tabular}{cc|ccccccccccc|ccc}
\hline
\multirow{2}{*}{Metric} &\multirow{2}{*}{Method} &\multicolumn{11}{c}{Synthetic Objects} &\multicolumn{3}{|c}{Real Objects} \\
& &FoldChair &Fridge &Laptop &Oven &Scissor &Stapler &USB 
&Washer&Blade &Storage &All & Fridge &Storage &All \\
\hline
\multirow{2}{*}{PSNR} 
&PARIS & 31.50 & \best{37.67} & \best{37.26} & 35.30 & \best{38.37} & 38.49 & 39.07 & \best{40.08} & 38.29 & 36.18 & 37.22 & 25.29 & \best{27.13} & 26.21 \\
&Ours & \best{34.46} & 37.11 & 34.09 & \best{37.06} & 38.29 & \best{39.13} & \best{39.64} & 38.50 & \best{41.16} & \best{37.24} & \best{37.67} & \best{27.05} & 25.38 & \best{26.22} \\
\hline
\multirow{2}{*}{SSIM} 
&PARIS & 0.985 & \best{0.994} & \best{0.991} & 0.980 & 0.996 & 0.995 & 0.992 & 0.991 & 0.996 & \best{0.993} & 0.991 & 0.898 & \best{0.953} & 0.926 \\
&Ours & \best{0.997} & 0.993 & 0.988 & \best{0.995} & \best{0.998} & \best{0.999} & \best{0.998} & \best{0.995} & \best{0.999} & 0.992 & \best{0.995} & \best{0.939} & 0.930 & \best{0.935} \\
\hline
\multirow{2}{*}{$\text{LPIPS}_{vgg}$} 
&PARIS & 0.045 & \best{0.032} & \best{0.020} & \best{0.045} & 0.015 & 0.019 & 0.029 & \best{0.029} & 0.017 & \best{0.095} & \best{0.035} & 0.188 & \best{0.139} & 0.164 \\
&Ours & \best{0.036} & 0.041 & 0.045 & 0.054 & \best{0.014} & \best{0.011} & \best{0.016} & 0.052 & \best{0.004} & 0.097 & 0.037 & \best{0.114} & 0.188 & \best{0.151} \\
\hline
\end{tabular}
}
}
\end{table*}

\begin{table*}[t]
\caption{{\textbf{Quantitative comparison for whole mesh reconstruction on PARIS data.} We report the average of CD-w over 10 trials. We bold \textbf{best} results on average of two states.}} 
\label{tab:app:exp_tsdf}
\renewcommand{\arraystretch}{1.2}
\resizebox{\linewidth}{!}{
{
\begin{tabular}{cc|ccccccccccc|ccc}
\hline
\multirow{2}{*}{Metric} &\multirow{2}{*}{Method} &\multicolumn{11}{c}{Synthetic Objects} &\multicolumn{3}{|c}{Real Objects} \\
& &FoldChair &Fridge &Laptop &Oven &Scissor &Stapler &USB 
&Washer&Blade &Storage &All & Fridge &Storage &All \\
\hline
\multirow{3}{*}{\shortstack{CD-w}}
&DTA
&\textbf{0.26} &0.70 &\textbf{0.34} &4.25 &\textbf{0.41} &\textbf{2.02} &\textbf{1.34} &\textbf{4.53} &\textbf{0.37} &\textbf{4.04} &\textbf{1.83} &2.13 &9.01 &5.57 \\
&TSDF with gt depth
& 0.30 & \textbf{0.56} & 0.47 & \textbf{3.60} & 0.49 & 2.78 & 1.60 & 5.73 & 0.54 & 5.13 & 2.12 & 3.15 & 131.86 & 67.51\\
&Ours
&0.36 &{0.59} &0.48 &{3.64} &0.68 &2.88 & 1.58 &6.05 & 0.63 & 5.17 &{2.21} &\textbf{1.37} &\textbf{2.84} &\textbf{2.11} 
\\
\hline
\end{tabular}
}
}
\end{table*}

{We provide additional comparisons with previous methods in this section. 
\paragraph{Scaled Axis Pos Metric.} Following DTA and PARIS, we multiply the 'Axis Pos' metric by 10 in \cref{tab:exp_2part} and \cref{tab:app:exp_2part}. While this metric shows minimal variation among current methods for synthetic objects, we also report the Axis Pos metric multiplied by 1000. As shown in \cref{tab:app:exp_ap}, \model demonstrates superior performance compared to DTA.
\paragraph{Perception-based Metrics.} To evaluate rendering quality, we assess perception-based metrics including LPIPS~\cite{zhang2018unreasonable}, SSIM~\cite{wang2004image}, and PSNR, with results shown in \cref{tab:app:exp_psnr}. While our primary focus aligns with previous methods on mesh reconstruction and articulation estimation, \model achieves comparable or superior performance relative to PARIS. 
\paragraph{Limited Improvement for CD-w on Simple Synthetic Objects.} Our method's performance on simple synthetic objects, particularly in terms of CD-w metric, is constrained by our use of TSDF for mesh extraction from Gaussian Splatting-rendered depths. To analyze this limitation, we compare against meshes reconstructed using ground-truth depth with TSDF. As shown in \cref{tab:app:exp_tsdf}, even with ground-truth depth input, TSDF-based reconstruction cannot surpass algorithms using marching cubes with NeRF, primarily due to the fundamental differences between TSDF and marching cubes algorithms on simple geometries. However, for complex or real-world objects where articulation reconstruction becomes more critical, the advantages of our model become evident. Additionally, TSDF with ground truth depth on real-world objects may produce poor-quality meshes (e.g., real\_storage) due to depth sensor noise, while our \model achieves high-quality reconstruction. Importantly, our primary objective is to create digital twins of real-world articulated objects, where \model demonstrates significant performance improvements, particularly for complex and real-world scenarios. 
}

\begin{figure}[t!]
 \centering
 \resizebox{\linewidth}{!}{\includegraphics[width=\linewidth]{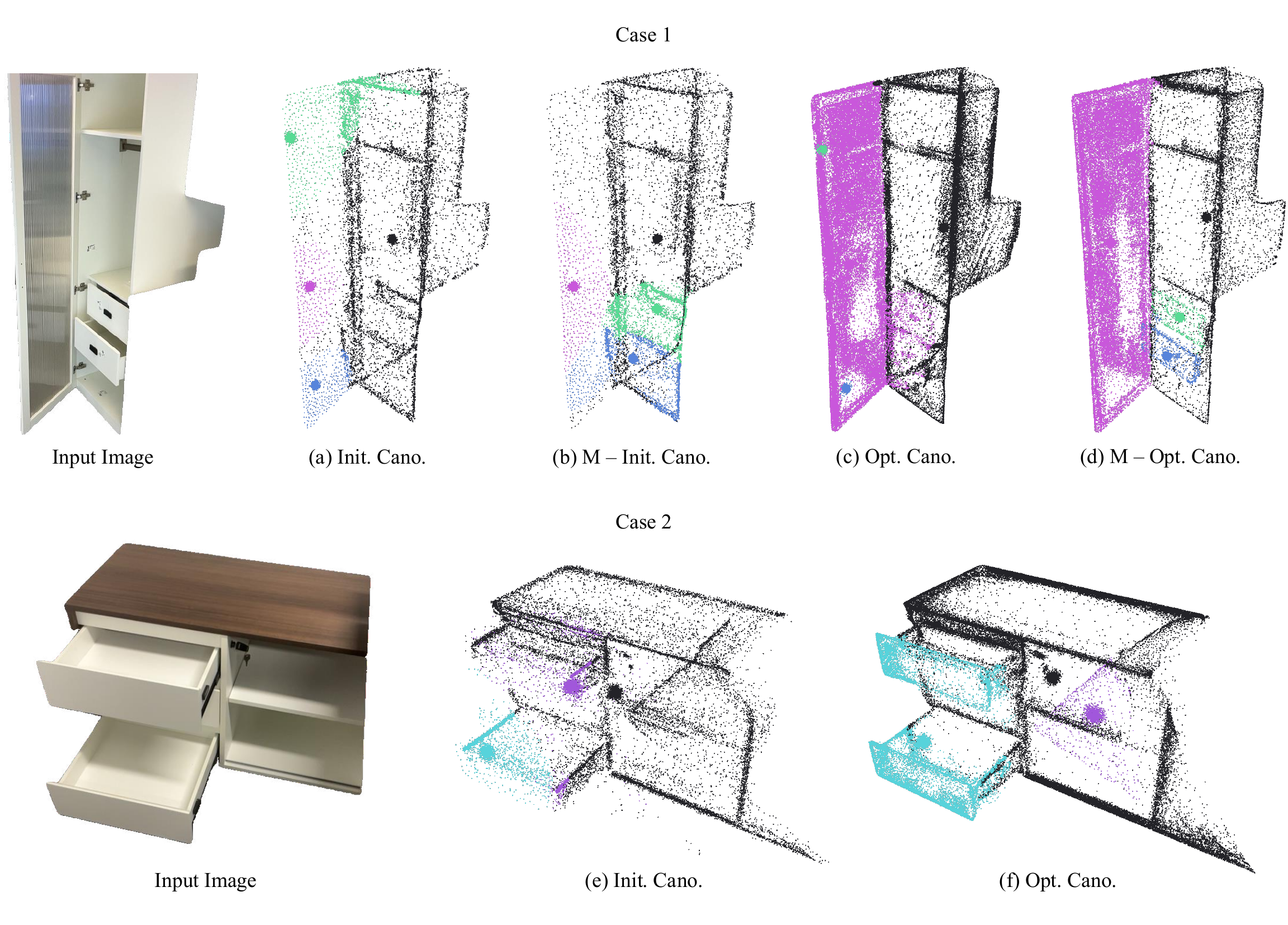}}
  \caption{{\textbf{Failure cases}. We illustrate failure cases of our \model. 'Init./Opt. Cano.' represents initialized and optimized Canonical Gaussians, while the prefix 'M' indicates manual correction of erroneous part centers.}}
 \label{fig:failure}
 \vspace{-10pt}
\end{figure}

{\subsection{Failure Cases}}
{\paragraph{Incorrect Initialization of Part Centers.} For real-world objects with multiple parts, clustering-derived part centers may be inaccurate (\cref{fig:failure} (a)) due to sensor noise, occlusion, and varying illumination conditions. These incorrectly initialized centers often persist through optimization, degrading performance for parts with misaligned centers (\cref{fig:failure} (c)). Manual correction of erroneous part centers prior to training (\cref{fig:failure} (b)) yields improved results (\cref{fig:failure} (d)). As discussed in \cref{app:limitation}, incorporating prior models like SAM for automatic, accurate part center initialization remains a promising direction for future work.}

{\paragraph{Similar Motions.} Our method exhibits limitations when handling parts with identical motion across states, as demonstrated in case 2 of \cref{fig:failure} where two drawers are pulled with the same distance. In such scenarios, the model tends to learn a single joint to fit both parts, failing to distinguish between the independently movable parts. As discussed in \cref{app:limitation}, expanding \model to incorporate additional states would provide richer motion information, potentially enabling better part separation.}

\begin{figure}[t!]
 \centering
 \resizebox{\linewidth}{!}{\includegraphics[width=\linewidth]{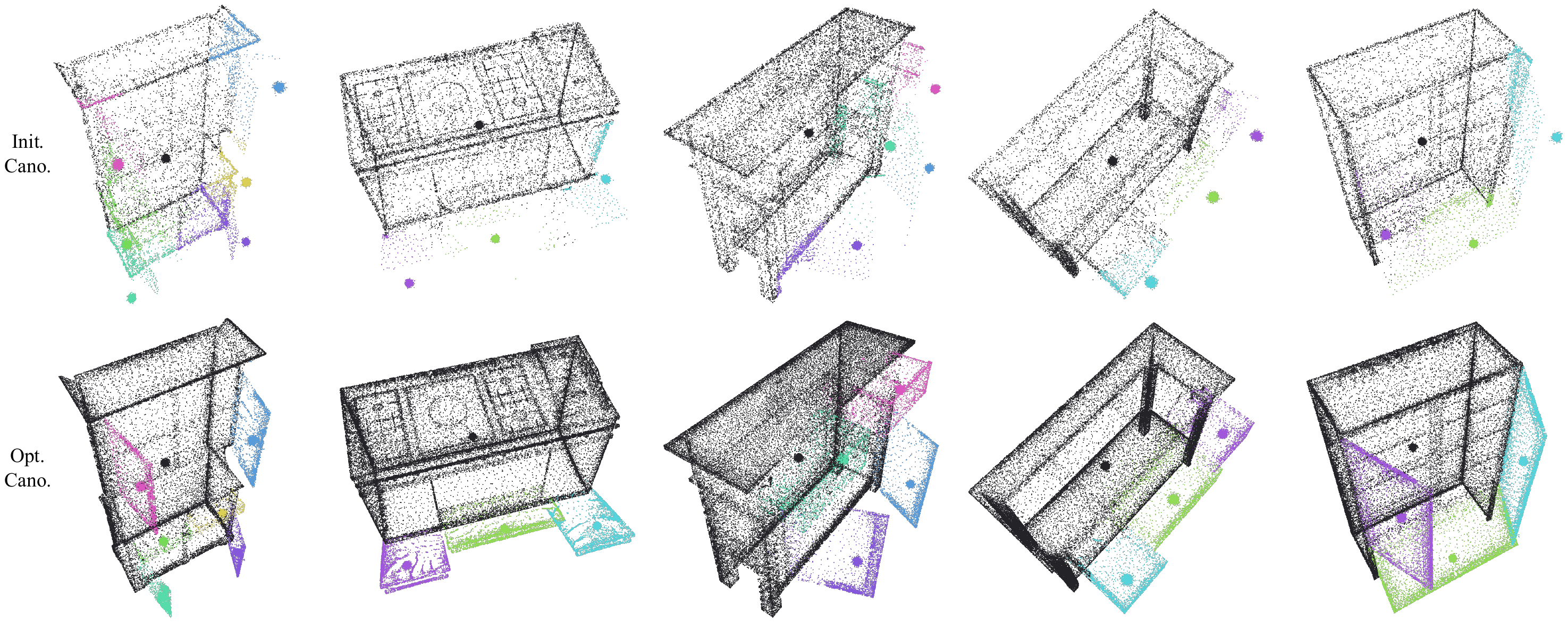}}
\caption{{\textbf{Evolution of canonical Gaussians}. We visualize the evolution of canonical Gaussians, showing both their part assignments and centers. Our initialization strategy begins with dense static Gaussians and sparse dynamic Gaussians. As training progresses, the Gaussians undergo densification while simultaneously refining their part centers and assignments. These visualization results demonstrate the effectiveness of \model.}}
 \label{fig:evolution}
 \vspace{-10pt}
\end{figure}

{\subsection{Evolution of Canonical Gaussians}}
{We visualize the evolution of canonical Gaussians in \cref{fig:evolution}, showing both their part assignments and centers. Our initialization strategy begins with dense static Gaussians and sparse dynamic Gaussians. As training progresses, the Gaussians undergo densification while simultaneously refining their part centers and assignments. These visualization results demonstrate the effectiveness of \model.}
{\subsection{Additional Qualitative Comparisons}}
{We provide additional qualitative comparisons on different datasets in the following pages.}
\begin{figure}[t!]
 \centering
 \resizebox{\linewidth}{!}{\includegraphics[width=\linewidth]{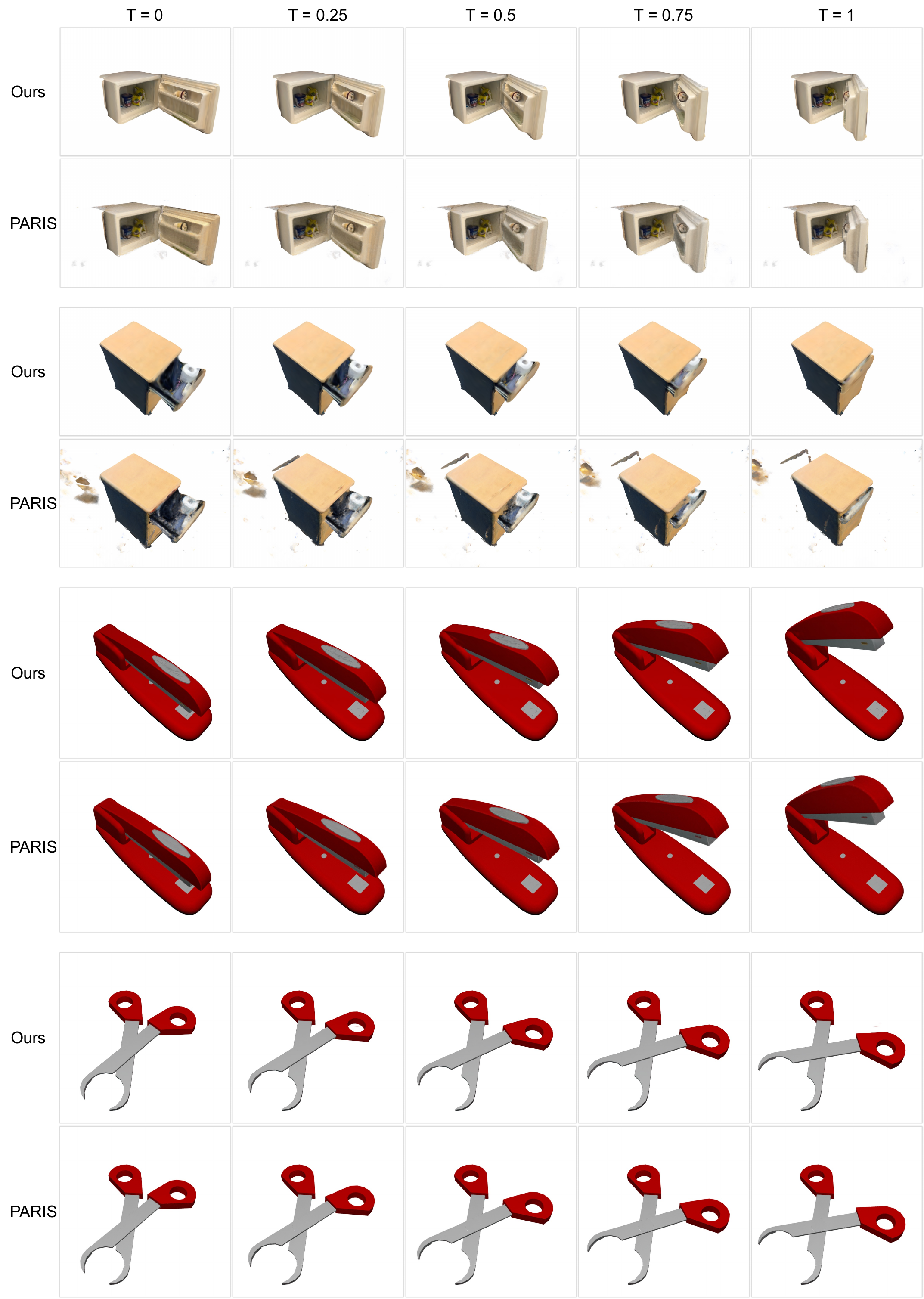}}
 \caption{{\textbf{Interpolation results on PARIS data.}}}
 \label{fig:interp_supp}
 \vspace{-10pt}
\end{figure}

\end{document}